# SPINEX: Similarity-based Predictions with Explainable Neighbors Exploration for Anomaly and Outlier Detection

M.Z. Naser[1,2], Ahmed Z. Naser[3]

[1]School of Civil & Environmental Engineering and Earth Sciences (SCEEES), Clemson University, USA
[2]Artificial Intelligence Research Institute for Science and Engineering (AIRISE), Clemson University, USA
E-mail: mznaser@clemson.edu, Website: www.mznaser.com
[3]Department of Mechanical Engineering, University of Manitoba, Canada, E-mail: a.naser@umanitoba.ca

**Abstract**
This paper presents a novel anomaly and outlier detection algorithm from the SPINEX (Similarity-based Predictions with Explainable Neighbors Exploration) family. This algorithm leverages the concept of similarity and higher-order interactions across multiple subspaces to identify outliers. A comprehensive set of experiments was conducted to evaluate the performance of SPINEX. This algorithm was examined against 21 commonly used anomaly detection algorithms, namely, namely, Angle-Based Outlier Detection (ABOD), Connectivity-Based Outlier Factor (COF), Copula-Based Outlier Detection (COPOD), ECOD, Elliptic Envelope (EE), Feature Bagging with KNN, Gaussian Mixture Models (GMM), Histogram-based Outlier Score (HBOS), Isolation Forest (IF), Isolation Neural Network Ensemble (INNE), Kernel Density Estimation (KDE), K-Nearest Neighbors (KNN), Lightweight Online Detector of Anomalies (LODA), Linear Model Deviation-based Detector (LMDD), Local Outlier Factor (LOF), Minimum Covariance Determinant (MCD), One-Class SVM (OCSVM), Quadratic MCD (QMCD), Robust Covariance (RC), Stochastic Outlier Selection (SOS), and Subspace Outlier Detection (SOD), and across 39 synthetic and real datasets from various domains and of a variety of dimensions and complexities. Furthermore, a complexity analysis was carried out to examine the complexity of the proposed algorithm. Our results demonstrate that SPINEX achieves superior performance, outperforms commonly used anomaly detection algorithms, and has moderate complexity (e.g., O(n log n × d)). More specifically, SPINEX was found to rank at the top of algorithms on the synthetic datasets and the 7$^{th}$ on the real datasets. Finally, a demonstration of the explainability capabilities of SPINEX, along with future research needs, is presented.

## 1.0 Introduction

Anomaly and outlier detection is a critical area of data mining and machine learning (ML) research, with wide-ranging applications across domains [1]. This particular area focuses on identifying data points or observations that deviate significantly from the expected patterns within a given data. These anomalies, also referred to as outliers, exceptions, peculiarities, or contaminants, often represent crucial information of interest to users [2]. Thus, accurately detecting anomalies can enhance system performance and reliability and may provide valuable insights that can lead to new discoveries or prevent potential failures. For example, detecting outliers in financial systems can help identify fraudulent transactions. Further, detecting the same in network security, healthcare, or industrial/engineering settings may identify intrusions, aid in the early detection of diseases, and predict equipment failures.

Historically, anomaly and outlier detection was handled via statistical methods that assumed a parametric model of normality, where any deviation from this model suggested an anomaly [3]. As data complexity grew, these traditional methods were seen to struggle. This paved the way for ML approaches as they can offer a more flexible framework to model and detect anomalies even in complex datasets where many variables interact in nonlinear and high-dimensional ways.

Anomaly and outlier detection techniques can be categorized based on the availability of labeled data [4]. For example, in supervised anomaly detection, models are trained on a labeled dataset containing both normal and anomalous samples. This method, while effective, is often limited by the availability of labeled anomalies. However, data obtained from most realistic scenarios do not contain labeled data; hence, such scenarios are more commonly impractical to label data. More recently, semi-supervised anomaly detection uses a small amount of labeled data alongside a larger pool of unlabeled data, providing a balance between the two approaches.

Outlier detection, while closely related to noise removal and accommodation, is considered a distinct field [5]. On the one hand, noise removal eliminates unwanted data elements that interfere with analysis. This process is typically a preprocessing step aimed at improving data quality before any substantive analysis begins. Noise in data can arise from various sources, such as measurement errors, data corruption, or limitations in data collection methods. The primary goal of noise removal is to enhance the signal-to-noise ratio, allowing the true patterns and relationships in the data to emerge more clearly. Rather than attempting to remove noise from the data, noise accommodation aims to develop robust statistical models in the presence of noise. This approach recognizes that eliminating noise may not always be possible or desirable but instead focuses on minimizing its impact on analytical outcomes [6].

The key difference between noise accommodation and outlier detection lies in their objectives. While noise accommodation seeks to mitigate the effect of anomalous data on statistical models, outlier detection aims to identify and often analyze these anomalous instances. In many cases, outliers are not merely noise to be ignored or accommodated but rather important signals that warrant further investigation. Another important aspect to consider is the potential value of outliers. While noise is generally considered detrimental to analysis, outliers can often provide valuable insights (e.g., they might point to new phenomena, challenge existing theories, or represent emerging trends or untapped opportunities).

As one can see, anomaly and outlier detection can be a challenging task. For example, defining normality in complex, high-dimensional, or dynamic datasets is non-trivial, as patterns may evolve temporally. The scarcity of labeled anomaly data often necessitates unsupervised or semi-supervised approaches. Real-world data noise can conceal true anomalies, which may demand robust algorithms to distinguish between genuine outliers and natural variations. Scalability is now crucial, given the availability of large datasets that require efficient processing. Additionally, contextual and collective anomalies introduce further complexity, as their detection requires consideration of broader data contexts or collective behaviors.

The concept of similarity lies at the heart of many anomaly detection techniques [7]. This fundamental idea theorizes that normal data points exhibit some degree of similarity to each other, while anomalies deviate significantly from this norm [8]. Thus, the concept of similarity serves as a powerful tool for distinguishing between regular patterns and outliers, and it continues to drive the basis for numerous detection algorithms. At its most basic level, similarity might be quantified as the distance between data points in a feature space [9]. Euclidean distance, for instance, is commonly used in many algorithms, particularly those dealing with numerical data. However, the concept of similarity extends far beyond simple geometric distance. For example, nearest neighbor approaches directly compare the similarity of a data point to its local neighborhood. Clustering techniques group similar data points together, with anomalies often appearing as singleton clusters or points far from cluster centers. Similarity is also embedded in the learned representations and reconstruction errors in more complex models like autoencoders [10].

The interest gained in similarity-based approaches lies in their intuitive nature and ability to capture complex, nonlinear relationships without requiring strong assumptions about the underlying distribution. This makes them particularly useful in high-dimensional spaces where traditional statistical methods may struggle. To leverage such a concept, we propose a novel algorithm, SPINEX, based on the concept of similarity and can offer explainable results. We provide a detailed description of our algorithm, including its mathematical formulation, implementation details, and empirical evaluation. We will demonstrate how our method performs on various benchmark datasets, comparing its performance against 21 state-of-the-art anomaly detection algorithms, commonly used anomaly and outlier detection algorithms, and across 39 synthetic and real datasets. Our experimental findings display the effectiveness and competitiveness of SPINEX compared to state-of-the-art algorithms and point out possible means for future improvements.

The rest of the paper is organized as follows: Section 2 describes SPINEX and explains each component in detail. Section 3 describes benchmarking algorithms and complexity analysis. Section 4 presents the selected datasets and our comparative results. The paper concludes with Sections 5 and 6 by presenting future research directions and our study's main findings.

## 2.0 Description of the SPINEX anomaly and outlier detection algorithm
This section describes the SPINEX in more detail herein.

### 2.1 General description
SPINEX is an unsupervised anomaly detection algorithm designed for high-dimensional data. SPINEX primarily utilizes similarity principles through distance metric calculations and feature interaction components. SPINEX measures the distances between data points using a specified metric—defaulting to Euclidean but configurable to others like Manhattan or Minkowski. Then, SPINEX assesses similarity directly, wherein data points closer to each other according to the selected distance metric are considered more similar, while those farther apart are deemed dissimilar. Further, SPINEX optionally computes both linear and nonlinear interactions between features. This method involves creating new features that represent the interactions (products) of pairs of original features. For linear interactions, the algorithm directly multiplies the values of two features, while for nonlinear interactions, it applies a transformation, such as logarithmic or square root, to the product depending on the nature of the data (e.g., ensuring non-negativity). These interactions provide a multi-dimensional view of data relationships, allowing the algorithm to capture more complex patterns and dependencies between features that are not apparent in the original data space.

By incorporating these interaction terms, SPINEX effectively increases the dimensional space in which the data points are compared, thereby enhancing the algorithm's sensitivity to subtle similarities or anomalies in the dataset. This is particularly useful in scenarios where anomalies are defined by unusual combinations of feature values rather than extreme individual feature values. Overall, SPINEX's proximity-based approach stems from the observation that anomalies are typically expected to manifest as points that deviate significantly from normal data points.

## *2.2 Detailed description*

A more detailed description of SPINEX's functions is provided herein.

### Initialization (__init__)

The __init__ method of the SPINEX algorithm initializes the anomaly detection framework, setting up the operational parameters and preparing the data for analysis. The method signature is as follows:

```
def __init__(self, data, column_names=None, use_weights=False, include_interactions=False,
distance_metric='euclidean',
        use_nonlinear=False, scaling_method=None, explainability_level='basic', anomaly_threshold=98,
        threshold_method='fixed', threshold_params=None, n_jobs=1):
```

This constructor takes multiple parameters that configure the behavior and functionality of the anomaly detection process:
- data (numpy.ndarray or pandas.DataFrame): The dataset to be analyzed. It can be provided as either a NumPy array or a pandas DataFrame.
- column_names (list of str, optional): Names for the columns if the data input is a NumPy array. Defaults to generating names like Feature1, Feature2, etc.
- use_weights (bool): If True, weights derived from the variance of each feature are used in distance calculations to emphasize more variable features.
- include_interactions (bool): If True, feature interactions (both linear and, if enabled, nonlinear) are computed to enrich the feature space.
- distance_metric (str): The metric used to calculate distances between data points for anomaly detection. Default is 'euclidean'.
- use_nonlinear (bool): Enables the computation of nonlinear interactions between features.
- scaling_method (str or None): Specifies the method used to scale the data. Options include standard, minmax, and robust.
- explainability_level (str): Determines the level of explainability in the outputs, ranging from basic to advanced.
- anomaly_threshold (int): The percentile used to determine the threshold for anomaly detection. Defaults to 98, indicating the 98$^{th}$ percentile.
- threshold_method (str): The method used to calculate the threshold for defining anomalies. Options include fixed, statistical, and adaptive_quantile.
- threshold_params (dict or None): Additional parameters for configuring the threshold method.
- n_jobs (int): The number of parallel jobs to run for distance calculations.

### Method: validate_and_convert_input(data, column_names)

This method serves as the entry point for data preprocessing in the SPINEX algorithm to ensure that input data is correctly formatted and labeled for subsequent analysis. The function accepts data in either a NumPy array or a pandas DataFrame format. If a NumPy array is provided, optional column names can be supplied; if none are given, default names are generated in the form of Feature1, Feature2, …, FeatureN for N columns.

Mathematically, let D be the input data matrix with dimensions m × n where m represents the number of samples and n the number of features. The function checks the type of D:

- If D is a NumPy array and column_names is None, it generates a list of names:
  
  column_names=["Feature"+str(i+1) for i in range(n)]
  
  and, converts D into a pandas DataFrame df:
  
  df=pd.DataFrame(D,columns=column_names)

- If D is already a pandas DataFrame, it is returned directly:
  
  return D

This ensures the data is always in a consistent DataFrame format for the operations that follow.

### Method: apply_scaling()

Scaling is crucial for normalizing feature magnitudes, especially when calculating distances. Thus, this method applies scaling based on the scaling_method parameter, which can be standard, minmax, or robust.

The scaling transformation is defined as follows:

- StandardScaler: Standardizes features by removing the mean and scaling to unit variance:
  
  $$x' = \frac{x - \mu}{\sigma}$$
  
  where μ and σ are the mean and standard deviation of the feature values, respectively.

- MinMaxScaler: Scales features to a given range, typically [0, 1]:
  
  $$x' = \frac{x - min(x)}{max(x) - min(x)}$$

- RobustScaler: Scales features using statistics that are robust to outliers:
  
  $$x' = \frac{x - Q_1(x)}{Q_3(x) - Q_1(x)}$$
  
  where $Q_1$ and $Q_3$ are the first and third quartiles, respectively.

### Method: precompute_interactions()

This method computes interactions between features as a means to enhance the feature space by considering both linear and potential nonlinear relationships. The number of interactions depends on whether nonlinear interactions are included.

Define the number of original features as n. The number of linear interactions is:

$$linear\_count = n(n - 1)/2$$

If nonlinear interactions are enabled, the number is doubled. Each linear interaction between features i and j is calculated as:

$$interaction_{i,j}^{nl} = x_i \times x_j$$

For nonlinear interactions, a transformation $f$ is applied:

$$interaction_{i,j}^{nl} = f(x_i \times x_j)$$

where $f$ could be a logarithmic or square root function, depending on the sign and magnitude of the multiplication.

### Method: calculate_feature_differences(row)

This method computes the distance between a given row r and all other rows in the dataset using the specified distance_metric. The distance metric d is applied as:

$$d(r, x) = \sqrt{\sum_{i=1}^{n} w_i (r_i - x_i)^2}$$

where, $w_i$ are the weights applied to the features, enhancing the importance of features with higher variances.

### Method: calculate_feature_differences_parallel(row)

This method is a straightforward parallelization wrapper for the calculate_feature_differences function. It allows SPINEX to leverage multi-core processing capabilities to enhance performance when calculating distances between data points in large datasets.

### Method: calculate_feature_contributions(row, baseline_differences)

This function computes the absolute differences between the features of a given row and a baseline, which is typically the mean or median of the dataset, to identify how much each feature contributes to the anomaly score.

For a dataset row r and a baseline vector b, the contribution of each feature is calculated as:

$$c_i = |r_i - b_i|$$

where $c_i$ is the contribution of the $i^{th}$ feature.

These contributions are summed to generate an overall anomaly score for each data point. This method is crucial in determining the significance of each feature's deviation from normal behavior, providing insights into the anomalous nature of the data points. The same is also elemental to explaining each of the predicted anomalies.

### Method: fixed_threshold(scores)

This method implements a fixed thresholding mechanism for anomaly detection based on percentile values. Given a set of anomaly scores s, the threshold T is determined as:

$$T = percentile(s, \tau)$$

where τ is the user-defined anomaly threshold percentile (e.g., 98).

This threshold is used to classify points as anomalies if their scores exceed *T*. The method ensures robustness by handling edge cases such as empty or *NaN-filled* score arrays for stable thresholding operation.

### Method: statistical_threshold(scores)

This method calculates an anomaly detection threshold based on the statistical properties of the score distribution. Given the mean $\mu$ and standard deviation $\sigma$ of the scores, the threshold *T* is computed as:

$$T = \mu + k\sigma$$

where, k is a multiplier provided by the user (default is 2). This method is effective in scenarios where the data distribution is assumed to be approximately normal, allowing for the capture of statistically significant outliers.

### Method: adaptive_quantile_threshold(scores)

The adaptive quantile threshold method dynamically adjusts the threshold based on the most recent data. This makes SPINEX particularly useful for streaming data or datasets with non-stationary distributions. For a scores vector s, the threshold *T* is calculated using:

$$T = percentile(s_{recent}, q \times 100)$$

where $s_{recent}$ is the subset of scores from the last nnn observations, and q is the quantile. This method adapts to changes in data behavior over time, providing a responsive and context-sensitive anomaly detection mechanism.

### Method: analyze()

The analyze method orchestrates the anomaly detection process. It starts by calculating feature differences for each row in parallel and then aggregates these differences to establish a baseline of normal behavior. Anomalies are identified by comparing individual behavior against this baseline using a statistical threshold method. The aggregation of differences for the baseline calculation involves computing the mean or median of the differences across all data points, providing a central tendency measure that serves as the anomaly detection threshold. Anomalies are those data points whose aggregated difference exceeds this threshold by a specified multiplier, typically set to capture the upper tail of the difference distribution.

### Method: get_predictions()

The get_predictions method in the SPINEX algorithm is the final step in the anomaly detection process. It generates predictions that classify each data point as either normal or an anomaly based on the anomaly scores and the threshold determined earlier in the analyze method. The method initializes a prediction array with a default value of 1, indicating normalcy for all data points in the dataset. It then examines the indices of the anomalies identified in the analyze method and updates the corresponding entries in the prediction array to -1, indicating anomalous points. Let p be the predictions vector, n be the number of observations in the dataset, and A be the set of indices corresponding to anomalies. The predictions are initially set as:

$$p = \{p_i = 1 \ \forall_i \in \{1, 2, \ldots, n\}\}$$

After identifying anomalies, the values in p at indices corresponding to anomalies are set to -1:

$$p_i = -1 \ \forall_i \in A$$

The output of the method is a tuple containing:
- predictions: an array where each element is either 1 (normal) or -1 (anomaly), corresponding to each data point in the dataset.
- self.anomaly_scores_all: the array of anomaly scores computed during the analysis, reflecting the degree of deviation of each data point from the expected normal behavior.

Two more functions exist: the visualize_transformations() and the visualize_feature_pairs(). These methods facilitate a visual examination of the relationships between pairs of features through scatter plots.

The complete class of SPINEX is shown below:

```
class SPINEX:
  def __init__(self, data, column_names=None, use_weights=False, include_interactions=False, distance_metric='euclidean',
          use_nonlinear=False, scaling_method=None, explainability_level='basic', anomaly_threshold=98,
          threshold_method='fixed', threshold_params=None, n_jobs=1):
    self.df = self.validate_and_convert_input(data, column_names)
    self.original_df = self.df.copy()  # Keep a copy of the original data for explainability
    self.use_weights = use_weights
    self.include_interactions = include_interactions
    self.distance_metric = distance_metric
    self.use_nonlinear = use_nonlinear
    self.scaling_method = scaling_method
    self.explainability_level = explainability_level
    self.anomaly_threshold = anomaly_threshold
    self.anomalies = None
    self.n_jobs = n_jobs  # Number of parallel jobs
    self.apply_scaling()
    if self.include_interactions:
       self.interactions = self.precompute_interactions()
    self.threshold_method = threshold_method
    self.threshold_params = threshold_params if threshold_params is not None else {}
    if self.use_weights:
      # Pre-calculate weights based on variance and store them
      self.weights = np.var(self.df.values, axis=0) + 1e-8  # Compute weights only once
      # Apply weights to the data
      self.weighted_df = self.df.values * np.sqrt(self.weights)
```

```python
@staticmethod
def validate_and_convert_input(data, column_names):
    if isinstance(data, np.ndarray):
        if column_names is None:
            column_names = [f'Feature{i+1}' for i in range(data.shape[1])]
        return pd.DataFrame(data, columns=column_names)
    elif not isinstance(data, pd.DataFrame):
        raise TypeError("Data should be a NumPy array or a pandas DataFrame")
    return data

def apply_scaling(self):
    if self.scaling_method is None:
        print("No scaling method provided; proceeding without scaling.")
        return

    scalers = {'standard': StandardScaler(), 'minmax': MinMaxScaler(), 'robust': RobustScaler()}
    scaler = scalers.get(self.scaling_method)
    if scaler:
        self.df[:] = scaler.fit_transform(self.df)
    else:
        valid_methods = ', '.join(scalers.keys())
        raise ValueError(f"Invalid scaling method: {self.scaling_method}. Valid options are: {valid_methods}")

def precompute_interactions(self):
    n_features = self.df.shape[1]
    columns = []

    # Adjust interaction count based on the presence of nonlinear interactions
    interaction_count = n_features * (n_features - 1) // 2
    if self.use_nonlinear:
        interaction_count *= 2  # Double the count for nonlinear interactions

    interactions = np.zeros((self.df.shape[0], interaction_count))

    k = 0
    for i in range(n_features):
        for j in range(i + 1, n_features):
            # Linear interaction
            interactions[:, k] = self.df.iloc[:, i] * self.df.iloc[:, j]
            columns.append(f'Interaction_{i+1}_{j+1}_linear')
            k += 1
            # Nonlinear interaction, only if enabled
            if self.use_nonlinear:
                interactions[:, k] = self.select_transformation(self.df.iloc[:, i] * self.df.iloc[:, j])
```

```python
            columns.append(f'Interaction_{i+1}_{j+1}_nonlinear')
            k += 1

    if len(columns) != interaction_count:
        raise ValueError("Mismatch between expected and actual number of columns")

    return pd.DataFrame(interactions, columns=columns)

def select_transformation(self, data):
    if np.any(data <= 0):
        return np.sqrt(np.abs(data))
    else:
        return np.log1p(data)

def calculate_feature_differences(self, row):
    if hasattr(self, 'weighted_df'):
        # If weights are used, calculate weighted distances
        weighted_row = row * np.sqrt(self.weights)
        distances = cdist(self.weighted_df, np.array([weighted_row]), metric=self.distance_metric)
    else:
        # If weights are not used, calculate distances using unweighted data
        distances = cdist(self.df.values, np.array([row]), metric=self.distance_metric)
    return distances.flatten()

def calculate_feature_differences_parallel(self, row):
    return self.calculate_feature_differences(row)

def calculate_feature_contributions(self, row, baseline_differences):
    contributions = {}
    for col_idx, col_name in enumerate(baseline_differences.index):
        contributions[col_name] = abs(row[col_name] - baseline_differences[col_name])
    return pd.Series(contributions)

def fixed_threshold(self, scores):
    # Check if the anomaly threshold is within the valid range
    if not 0 <= self.anomaly_threshold <= 100:
        raise ValueError("Anomaly threshold must be between 0 and 100")

    # Check if the scores array is empty
    if scores.size == 0:
        raise ValueError("Scores array is empty.")

    # Remove NaN values from scores to ensure percentile calculation is valid
    if np.isnan(scores).any():
        scores = scores[~np.isnan(scores)]
```

```python
        if scores.size == 0:  # Check again in case all were NaN
            raise ValueError("Scores array contains only NaN values.")

    return np.percentile(scores, self.anomaly_threshold)

def statistical_threshold(self, scores):
    mean = scores.mean()
    std = scores.std()
    multiplier = self.threshold_params.get('multiplier', 2)  # Default multiplier is 2
    return mean + multiplier * std

def adaptive_quantile_threshold(self, scores):
    window_size = self.threshold_params.get('window_size', 50)  # Default window size
    quantile = self.threshold_params.get('quantile', 0.95)  # Default quantile
    if len(scores) <= window_size:
        return np.percentile(scores, quantile * 100)
    else:
        recent_scores = scores[-window_size:]
        return np.percentile(recent_scores, quantile * 100)

def analyze(self):
    # Calculate feature differences for all rows in df using parallel processing
    feature_differences = Parallel(n_jobs=self.n_jobs)(delayed(self.calculate_feature_differences_parallel)(row.values) for _, row in self.df.iterrows())

    feature_differences = np.array(feature_differences)
    if feature_differences.ndim == 2:
        feature_differences_df = pd.DataFrame(feature_differences, columns=[f'Feature_{i}' for i in range(feature_differences.shape[1])])

        # Compute baseline differences as the mean of differences
        baseline_differences = feature_differences_df.mean()

        # Calculate feature contributions
        feature_contributions = feature_differences_df.apply(lambda row: abs(row - baseline_differences), axis=1)

        # Calculate and store anomaly scores in the class
        self.anomaly_scores = feature_contributions.sum(axis=1)  # Now storing anomaly scores as an attribute
        self.anomaly_scores_all = self.anomaly_scores.copy()

        # Select threshold based on the method
        if self.threshold_method == 'fixed':
            threshold = self.fixed_threshold(self.anomaly_scores)
        elif self.threshold_method == 'statistical':
```

```python
            threshold = self.statistical_threshold(self.anomaly_scores)
        elif self.threshold_method == 'adaptive_quantile':
            threshold = self.adaptive_quantile_threshold(self.anomaly_scores)
        else:
            raise ValueError(f"Unknown threshold method: {self.threshold_method}")

        # Identify anomalies based on the threshold
        self.anomalies = self.original_df[self.anomaly_scores > threshold]

        if self.explainability_level == 'advanced':
            self.visualize_transformations()
            if not self.anomalies.empty:
                self.visualize_feature_pairs()
    else:
        raise ValueError("Feature differences did not return valid data.")

def get_predictions(self):
    predictions = np.ones(len(self.df), dtype=int)  # Initialize all as normal (1)
    if self.anomalies is not None:
        anomaly_indices = self.anomalies.index
        predictions[anomaly_indices] = -1  # Set anomalies to -1
    return predictions, self.anomaly_scores_all

def visualize_transformations(self):
    fig, axes = plt.subplots(1, 2, figsize=(12, 6))
    sns.kdeplot(data=self.original_df, ax=axes[0], fill=True)
    axes[0].set_title('Original Data Distribution')
    sns.kdeplot(data=self.df, ax=axes[1], fill=True)
    axes[1].set_title('Transformed Data Distribution')
    plt.show()

def visualize_feature_pairs(self):
    feature_names = self.df.columns
    n_features = len(feature_names)
    if n_features < 2:
        raise ValueError("Not enough features for pair-wise visualization")
    plt.figure(figsize=(15, 15))
    plot_number = 1
    for i in range(n_features):
        for j in range(i + 1, n_features):
            plt.subplot(n_features - 1, n_features - 1, plot_number)
            plt.scatter(self.df[feature_names[i]], self.df[feature_names[j]], alpha=0.5, label='Normal Data' if plot_number == 1 else "")
            plt.scatter(self.anomalies[feature_names[i]], self.anomalies[feature_names[j]], color='red', marker='s', label='Anomalies' if plot_number == 1 else "")
```

```
        plt.xlabel(feature_names[i])
        plt.ylabel(feature_names[j])
        if plot_number == 1:
            plt.legend()
        plot_number += 1
plt.suptitle('Anomaly Visualization Across Feature Pairs')
plt.tight_layout()
plt.show()
```

## 3.0 Description of benchmarking algorithms, experiments, and functions

This section describes the experimental examination used to benchmark SPINEX. For a start, SPINEX was examined against 21 commonly used optimization algorithms, namely, Angle-Based Outlier Detection (ABOD), Connectivity-Based Outlier Factor (COF), Copula-Based Outlier Detection (COPOD), ECOD, Elliptic Envelope (EE), Feature Bagging with KNN, Gaussian Mixture Models (GMM), Histogram-based Outlier Score (HBOS), Isolation Forest (IF), Isolation Neural Network Ensemble (INNE), Kernel Density Estimation (KDE), K-Nearest Neighbors (KNN), Lightweight Online Detector of Anomalies (LODA), Linear Model Deviation-based Detector (LMDD), Local Outlier Factor (LOF), Minimum Covariance Determinant (MCD), One-Class SVM (OCSVM), Quadratic MCD (QMCD), Robust Covariance (RC), Stochastic Outlier Selection (SOS), and Subspace Outlier Detection (SOD). A brief description of each of the utilized algorithms is presented herein, and we invite our readers to cross examine the sources for these algorithms for additional details. Our analysis utilized all of these algorithms in default settings. Table 1 compares these algorithms.

*3.1 Angle-Based Outlier Detection (ABOD):*

The Angle-Based Outlier Detection (ABOD) was developed by Kriegel et al. in 2008 [11] to address the challenges of anomaly detection in high-dimensional spaces where traditional distance-based methods often struggle. The ABOD builds on the observation that for normal points, the angles formed with pairs of other points tend to vary widely, while for outliers, these angles tend to be consistently small. As such, the algorithm computes an angle-based outlier factor for each point, which is the variance of the angles formed by the point with pairs of other points in the dataset. Points with small variances in these angles are considered potential outliers. ABOD's key advantage is its effectiveness in high-dimensional spaces, as it does not rely directly on distance. Additionally, ABOD does not require explicit parameters like the number of neighbors, and it can detect both global and local outliers. However, the naive implementation of ABOD has high computational complexity, which can be prohibitive for large datasets, although approximation methods like FastABOD can mitigate this issue [12].

*3.2 Connectivity-Based Outlier Factor (COF):*

The Connectivity-Based Outlier Factor (COF) is an extension of the Local Outlier Factor (LOF) algorithm that was proposed by Tang et al. in 2002 [13]. The COF introduces the concept of a set-based nearest path (SBN-path) to measure the connectivity between points, replacing the Euclidean distance used in LOF. Thus, for each point, COF calculates an average chaining distance based on these SBN-paths, then computes the COF score as the ratio of the average chaining distances of a point's neighbors to its own average chaining distance. Points with COF values significantly larger than 1 are considered potential anomalies. This approach allows COF to better handle datasets where clusters have different densities and to more effectively detect outliers in elongated clusters. However, like LOF, COF's performance can also be sensitive to the choice of the number of nearest neighbors.

*3.3 Copula-Based Outlier Detection (COPOD):*

COPOD is a parameter-free probabilistic approach to detecting outliers in multivariate data, proposed by Li et al. in 2020 [14]. COPOD is a non-parametric method that models the dependency structure between dimensions of data using copulas. This algorithm can effectively capture the tail dependencies and detect anomalies in complex multivariate distributions without the assumption of normality. The outlier score is based on the empirical copula of observed data points.

*3.4 Elliptic Envelope (EE):*

The Elliptic Envelope method is an object for detecting outliers in Gaussian distributed data using a robust covariance estimation [15]. It assumes that the data is generated from a single Gaussian distribution and fits an ellipse to the central data points, ignoring points outside a specified contamination parameter. Thus, points outside the ellipse defined by several standard deviations are considered outliers.

*3.5 Empirical Cumulative Distribution Function-based Outlier Detection (ECOD):*

The Empirical Cumulative Distribution Function-based Outlier Detection (ECOD) is a parameter-free method that utilizes empirical cumulative distribution functions to identify anomalies (in both univariate and multivariate data). ECOD begins by independently computing the empirical cumulative distribution function for each feature. Then, this algorithm transforms the data to the [0,1] space using these functions. The outlier score for each point is computed as the maximum absolute difference between its transformed values and 0.5 across all features. ECOD offers several advantages, including computational efficiency with a time complexity of *O(n log n)*, robustness to data dimensionality, and interpretable probabilistic scores. As ECOD assumes feature independence, this algorithm may not capture complex feature interactions as effectively.

*3.6 Feature Bagging with KNN*

Feature Bagging with KNN is an ensemble method for anomaly detection that combines multiple base KNN detectors to improve robustness and performance [16]. This algorithm creates multiple subsets of features by randomly selecting a portion of the original features. For each subset, it trains a KNN anomaly detector and calculates anomaly scores for each data point. The final anomaly score for a point is then determined by combining the scores from all detectors, typically through averaging or taking the maximum. This approach addresses some limitations of standard KNN anomaly detection, such as sensitivity to irrelevant features and high dimensionality. Still, such improvement comes at the cost of increased computational complexity (compared to standard KNN) and can be sensitive to the number of feature subsets and subset size. The method is particularly useful in scenarios with high-dimensional data or when there is uncertainty about which features are most relevant for anomaly detection.

*3.7 Gaussian Mixture Models (GMM)*

The GMM for anomaly detection is based on fitting a mixture of Gaussian distributions to the data and then identifying points with low likelihood under this model. Algorithmic parameters are typically estimated using the Expectation-Maximization (EM) algorithm. Once the model is fitted, anomalies are identified as points with low likelihood under the model or, equivalently, high negative log likelihood. GMM has the advantage of being able to model complex, multimodal distributions and provide probabilistic anomaly scores. However, specifying the number of mixture components in advance can be challenging. Additionally, GMM can be sensitive to initialization and may converge to local optima.

### 3.8 Histogram-based Outlier Score (HBOS):

The Histogram-based Outlier Score (HBOS) was developed by Goldstein and Dengel in 2012 [17]. The HBOS can be described as a statistical anomaly detection method designed for fast processing of large datasets. HBOS independently constructs histograms for each feature and then combines information from these histograms to compute an anomaly score. Then, the algorithm calculates bin heights as normalized frequencies for each feature and then computes the score for each data point as the sum of the logarithms of inverse bin heights across all features. Naturally, higher scores indicate a higher likelihood of the point being an anomaly. HBOS offers several advantages, including linear time complexity (i.e., $O(n)$), ease of interpretation, and the ability to handle mixed data types. On the other hand, this algorithm assumes feature independence, which may not always hold true and can be sensitive to the choice of bin width. The method may also face challenges in high-dimensional spaces where the likelihood of a point being in a low-density region by chance increases. HBOS can be particularly useful when computational efficiency is crucial and where the independence assumption is reasonable [18].

### 3.9 Isolation Forest (IF):

The Isolation Forest (IF) was Liu et al. in 2008 [19]. This algorithm is based on the principle that anomalies are few, rare, and different; hence, they can be easily isolated in a dataset. The IF algorithm constructs a forest of random decision trees, called isolation trees, and identifies anomalies as instances with short average path lengths on these trees. For example, the algorithm recursively partitions the data in each tree by randomly selecting a feature and a split value, continuing until each data point is isolated or a specified tree height is reached. Then, an anomaly score for a point is calculated based on the average path length across all trees, with shorter paths indicating a higher likelihood of being an anomaly (i.e., anomalies are instances with scores close to 1, while normal points have scores much smaller than 0.5). The IF algorithm performs well in high-dimensional datasets and is particularly effective when anomalies are scattered and in small numbers. Its time complexity is linear with the number of samples and features, and it can be efficient for large datasets. However, IF may struggle with datasets where anomalies form clusters or when normal data is not homogeneous [20]. The contamination parameter in the IF implementation determines the proportion of outliers in the dataset, which, in turn, influences the threshold for anomaly classification.

### 3.10 Isolation Neural Network Ensemble (INNE)

INNE combines the concept of isolation with nearest neighbor methods to enhance detection in high-dimensional spaces. Thus, the INNE is an ensemble method for anomaly detection that combines multiple neural networks trained to isolate individual data points. Each neural network in the ensemble is trained on a random subset of features and a random subset of the data. The networks learn to separate individual points from the rest of the data, similar to the concept of Isolation Forests. The anomaly score for a point is then computed as the average path length (or a similar metric) across all networks in the ensemble. Points that are easier to isolate (i.e., have shorter average path lengths) are considered potential anomalies. INNE has the advantage of effectively capturing complex, nonlinear relationships in the data and handling high-dimensional datasets. The ensemble approach also provides robustness against noise and irrelevant features. However, INNE can be computationally intensive, especially for large datasets or when using a large number of networks in the ensemble.

### 3.11 Kernel Density Estimation (KDE)

The KDE is a non-parametric method for estimating the probability density function of a random variable based on a finite data sample [21]. This algorithm can be used to identify data points that lie in regions of low density. The algorithm estimates the density at each point using a kernel function centered at that point, typically a Gaussian kernel. Points with low density estimates are considered potential anomalies. The bandwidth parameter $h$ controls the smoothness of the density estimate and can significantly affect the results. KDE has the advantage of being able to capture complex, multimodal distributions without assuming a specific parametric form. However, it can be computationally expensive for large datasets and may struggle in high-dimensional spaces.

### 3.12 K-Nearest Neighbors (KNN) for Anomaly Detection

When adapted for anomaly detection, the K-Nearest Neighbors (KNN) algorithm can be best described as a distance-based method that relies on the principle that normal data points have close neighbors while anomalies are located far from their closest neighbors. Thus, KNN calculates the average distance to its $k$ nearest neighbors for each data point as an anomaly score. Points with higher scores are considered potential anomalies. The algorithm is intuitive and easy to implement, capable of working well in various scenarios, including when normal data forms multiple clusters. However, this algorithm may face challenges such as high computational complexity for large datasets and dependence on the choice of $k$. This algorithm has a similar time complexity to the LOF algorithm.

### 3.13 Lightweight Online Detector of Anomalies (LODA)

The LODA algorithm was developed by Pevny [22] as a fast, online anomaly detection algorithm designed for streaming data and large datasets. LODA works by projecting the data onto multiple random one-dimensional subspaces and constructing histograms in these subspaces. Each projection maintains a histogram of the projected values, and the anomaly score for a new point is computed as the negative log-likelihood of its projected values across all histograms. LODA can be computationally efficient with linear time complexity, handle high-dimensional data, and be updated incrementally. Additionally, LODA provides interpretable results by identifying which features contribute most to a point's anomaly score. However, LODA may not capture complex, multi-dimensional relationships in the data as effectively as some more sophisticated methods.

### 3.14 Linear Median Deviation-based Detector (LMDD)

LMDD is a robust statistic that seeks to minimize the squared deviations from the median of the dataset. The algorithm first fits a model to the dataset, and then it computes the residuals for each data point (i.e., which represents the deviation from the model). The magnitude of these residuals serves as an anomaly score, with larger residuals indicating potential outliers. One advantage of LMDD is its simplicity and interpretability. However, it can be sensitive to outliers in the model fitting stage, potentially masking some anomalies.

### 3.15 Local Outlier Factor (LOF)

The Local Outlier Factor (LOF) algorithm was developed by Breunig et al. in 2000 [23] as a density-based method for detecting anomalies. Unlike global outlier detection methods, LOF identifies anomalies by comparing a point's local density to its neighbors' local densities. Simply, the LOF considers the samples that have a substantially lower density than their neighbors as outliers. This approach allows LOF to detect local anomalies that might be missed by global methods – notably in datasets with varying densities. The LOF starts by calculating a local reachability density for each point based on its *k* nearest neighbors, then computes the LOF score as the ratio of the average local reachability density of a point's neighbors to its own local reachability density. Points with LOF scores significantly larger than 1 are considered outliers. LOF's strength lies in its ability to detect outliers in datasets with varying densities and its robustness to different distance measures. However, it can be computationally intensive for large datasets (as it has a time complexity equal to a time complexity of $O(n^2)$) and sensitive to the choice of the *k* parameter [24].

*3.16 Minimum Covariance Determinant (MCD)*

The Minimum Covariance Determinant (MCD) was proposed by Rousseeuw [25]. The MCD is a highly robust estimator of multivariate location and scatter and is designed to resist the effect of outliers in the data. The algorithm aims to find the subset of *h* observations (out of *n* total observations) whose covariance matrix has the lowest determinant. Typically, *h* is chosen to be about 75% of *n* to balance robustness and efficiency. The MCD algorithm iteratively selects subsets of the data, computes their mean and covariance, and keeps the subset with the lowest covariance determinant. Once the optimal subset is found, it is used to compute the final estimates of location and scatter. These robust estimates can then be used to compute Mahalanobis distances for all points, with large distances indicating potential outliers. The MCD method is particularly useful in multivariate settings where outliers might not be apparent in any single dimension but become visible in the multivariate structure of the data. While the original MCD can be computationally intensive for large datasets, fast approximation algorithms have been developed to make it more practical [26].

*3.17 One Class Support Vector Machine (OCSVM)*

The One-Class Support Vector Machine (OCSVM) is an extension of the Support Vector Machine algorithm designed for unsupervised anomaly detection. This algorithm was proposed by Schölkopf et al. [27] to learn a decision boundary that encapsulates the majority of the data points. Then, any point that falls outside this boundary is treated as an outlier. More specifically, OCSVM maps input data into a high-dimensional feature space via a kernel function and then seeks to find the maximal margin hyperplane that separates the data from the origin. This algorithm solves an optimization problem to determine this hyperplane, with the decision function classifying new points based on their position relative to the hyperplane. OCSVM is particularly effective when the normal data is well-clustered and separable from anomalies in the feature space. However, OCSVM can be sensitive to the choice of kernel and its parameters, and it may struggle with datasets containing multiple normal clusters or when anomalies are not well-separated from normal data. OCSVM can also be computationally intensive for large datasets and has a time complexity of $O(n^2)$ to $O(n^3)$, where *n* is the number of training samples.

*3.18 Quantile Minimum Covariance Determinant (QMCD)*

QMCD is an extension of the Minimum Covariance Determinant (MCD) method that incorporates quadratic terms to better handle nonlinear relationships in the data [28]. QMCD aims to provide a more robust and flexible approach to outlier detection.

*3.19 Robust Covariance (RC)*

Robust Covariance is an outlier detection method that uses a robust estimator of covariance to identify anomalies in multivariate data [29]. This method is similar to the Elliptic Envelope method but provides more flexibility in the choice of the robust covariance estimator.

*3.20 Stochastic Outlier Selection (SOS)*

The Stochastic Outlier Selection (SOS) is an unsupervised anomaly detection algorithm proposed by Janssens et al. in 2012 [30]. This algorithm is based on the concept of affinity between data points and uses a probabilistic approach to identify outliers. The algorithm begins by computing pairwise distances between all points in the dataset. These distances are then converted to affinities using a Gaussian kernel, where the bandwidth of the kernel is determined adaptively for each point based on a user-specified perplexity value. The affinities are normalized to obtain binding probabilities that represent the probability that a point would choose another specific point as its neighbor. The outlier probability for each point is then computed as the product of the probability that it would not be chosen as a neighbor by any other point. This approach allows SOS to capture complex, nonlinear distributions and relationships in the data and adapt to varying densities. One of the key advantages of SOS is that it provides a probabilistic outlier score, which can be more interpretable than distance-based scores. However, SOS can be computationally expensive for large datasets due to the need to compute all pairwise distances.

*3.21 Subspace Outlier Detection (SOD)*

The Subspace Outlier Detection (SOD) algorithm was proposed by Kriegel et al. in 2009 [31]. SOD is designed to detect outliers in high-dimensional spaces by considering relevant subspaces for each data point. For example, for each point, SOD first identifies its $k$ nearest neighbors, then computes a subspace spanned by these neighbors using techniques like principal component analysis [32]. The point and its neighbors are then projected onto this subspace, and the distance of the point to the centroid of its neighbors in this subspace is computed. This distance is normalized by the spread of the neighbors in the subspace to produce an outlier score. Points with high normalized distances are considered outliers. The strength of this algorithm lies in its ability to detect outliers that are only visible in specific subspaces, making it particularly effective in high-dimensional datasets where outliers might be obscured when considering all dimensions simultaneously. The algorithm adapts to local data characteristics, providing a flexible approach to outlier detection. However, SOD can be sensitive to the choice of the number of neighbors and can be computationally expensive for large datasets.

*3.22 Complexity analysis*

A basic comparative complexity analysis was conducted between the above algorithms and SPINEX. This analysis was carried out across 100, 1000, and 10000 samples (n) with 50, 100, 500, and 1000 features (d) – see Figure 1 and Table 1. This analysis shows that SPINEX, with an empirical complexity of $O(n \log n \times d)$, demonstrates competitive performance in terms of computational complexity compared to other anomaly detection algorithms. This complexity places SPINEX in the middle tier of efficiency among the examined algorithms, which have better scalability than quadratic algorithms. More specifically, our findings categorize them into three broad groups based on their complexity:

1. Linear or Near-Linear Complexity: Algorithms like HBOS, IF, and LODA, with O(n × d) complexity, represent the most efficient category. These algorithms scale linearly with both the number of samples and features, which can be preferred for large datasets. SPINEX is only marginally slower than the aforementioned algorithms due to its additional logarithmic factor.
2. Log-Linear Complexity: SPINEX falls into this category along with algorithms like ECOD, FeatureBagging, INNE, LOF, and ABOD. These algorithms, with complexities around O(n log n × d) or O(n log n × d log d), offer a good balance between computational efficiency and detection capability, with only a slight penalty compared to linear algorithms.
3. Higher Complexity: Algorithms such as KDE, LMDD, OCSVM, QMCD, and SOS, with complexities of $O(n^{1.5} \times d)$ or higher, represent the most computationally intensive methods.

It is important to note that computational complexity is just one factor in algorithm selection. Detection accuracy, robustness to different data distributions, interpretability of results, and specific use case requirements are all crucial considerations. SPINEX's moderate complexity suggests it might balance computational efficiency and detection capability well.

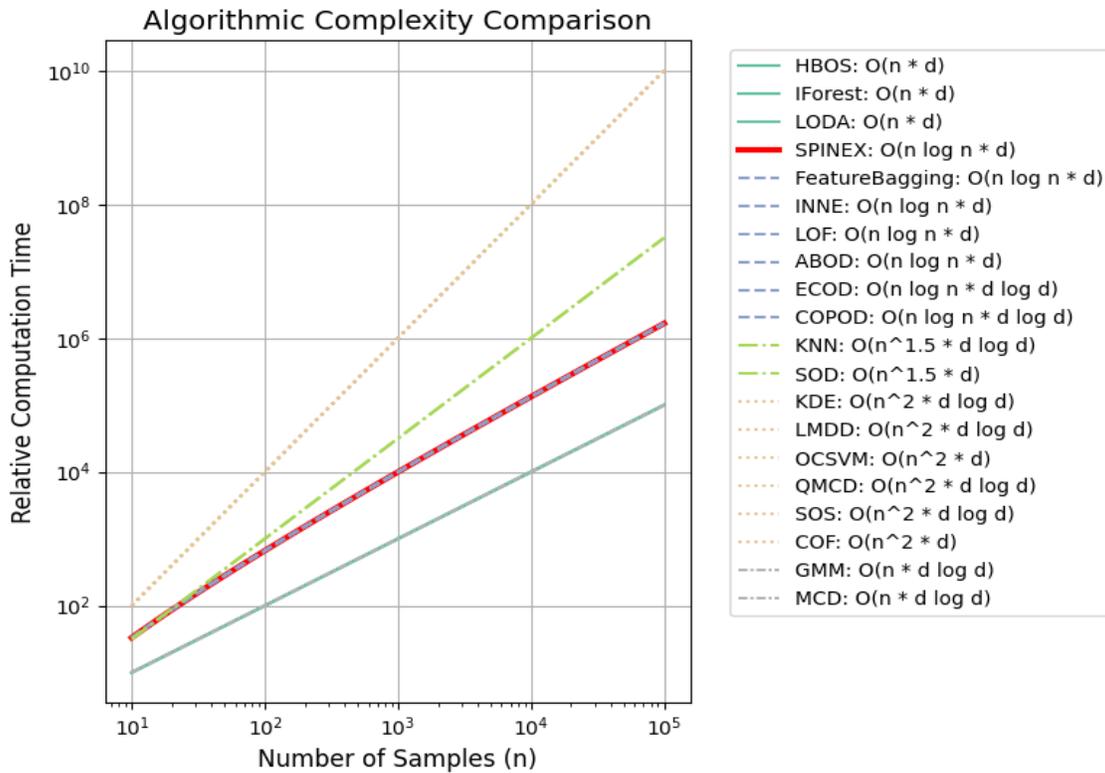

Fig. 1 Complexity analysis

269  Table 1 A comparison between anomaly algorithms

| Algorithm | Type | Strengths | Weaknesses | Computational complexity* |
|---|---|---|---|---|
| Angle-Based Outlier Detection (ABOD) | Angle-based | Effective in high-dimensional spaces; Less affected by data density. | Computationally intensive; Performance can vary with dimensionality. | $O(n \log n \times d)$ |
| Connectivity-Based Outlier Factor (COF) | Distance-based | Good for datasets with varying densities; Detailed local outlier detection. | High computational complexity; Not suitable for large datasets. | $O(n^2 \times d)$ |
| Copula-Based Outlier Detection (COPOD) | Copula-based | No parameter tuning needed; Effective with different data distributions. | Less interpretable; May struggle with very high-dimensional data. | $O(n \log n \times d \log d)$ |
| ECOD | Empirical Distribution | Simple and efficient; No assumptions on data distribution. | May be less effective for complex or highly noisy data. | $O(n \log n * d \log d)$ |
| Elliptic Envelope (EE) | Covariance-based | Effective for Gaussian distributed data; Intuitive model assumption. | Poor performance with non-Gaussian data; Sensitive to outliers. | - |
| Feature Bagging with KNN | Ensemble | Combines multiple models for robustness; Reduces risk of overfitting. | Increased computational cost; Depends on the effectiveness of individual models. | $O(n \log n \times d)$ |
| Gaussian Mixture Models (GMM) | Model-based | Flexibility in modeling data distributions; Effective in many scenarios. | Requires assumptions on component number; Can be sensitive to initialization. | $O(n \times d \log d)$ |
| Histogram-based Outlier Score (HBOS) | Histogram-based | Fast and scalable; Good for large datasets. | Assumes feature independence; Poor performance with correlated features. | $O(n \times d)$ |
| Isolation Forest (IF) | Tree-based | Efficient with high-dimensional data; Handles large datasets well. | Less effective with very small datasets; Performance can degrade with noisy data. | $O(n \times d)$ |
| Isolation Neural Network Ensemble (INNE) | Neural Network | Robust performance; Combines multiple neural networks for diversity. | Requires large data for training; Computationally expensive. | $O(n \log n \times d)$ |
| Kernel Density Estimation (KDE) | Density-based | Good for data with underlying structures; Flexible kernel choice. | Sensitive to bandwidth parameter; Struggles with high dimensions. | $O(n^2 \times d \log d)$ |
| K-Nearest Neighbors (KNN) | Distance-based | Intuitive and simple; Effective in identifying global outliers. | Computationally expensive; Not ideal for high-dimensional data. | $O(n^{1.5} \times d \log d)$ |
| Lightweight Online Detector of | Histogram-based | Lightweight and suitable for online environments; Fast and scalable. | Performance can be limited by histogram binning; Less effective with complex patterns. | $O(n \times d)$ |
| Anomalies (LODA) | | | | |
| Linear Model Deviation-based Detector (LMDD) | Linear models | Simple and fast for data with linear relations. | Ineffective with nonlinear or complex patterns. | $O(n^2 \times d \log d)$ |
| Local Outlier Factor (LOF) | Density-based | Detects local outliers effectively; Adaptable to varying densities. | Not suited for high-dimensional data; Sensitive to parameter k. | $O(n \log n \times d)$ |
| Minimum Covariance Determinant (MCD) | Robust Statistics | Resistant to outliers; Effective in moderate-dimensional datasets. | Computationally intensive; Not scalable to very large datasets. | $O(n \times d \log d)$ |
| One-Class SVM (OCSVM) | Support Vector Machine | Effective in high-dimensional space; Good for novelty detection. | Sensitive to parameter settings; Can be computationally intensive. | $O(n^2 \times d)$ |
| Quadratic MCD (QMCD) | Robust Statistics | Enhances MCD by better handling data shape; Robust against outliers. | Higher computational demand than MCD; Scalability issues. | $O(n^2 \times d \log d)$ |
| Robust Covariance (RC) | Covariance-based | Effective for multivariate data; Robust to outliers. | Requires assumption of Gaussian distribution; Computationally heavy. | - |
| SPINEX | Similarity-based + Feature Interaction | Effective for multivariate data; Robust to outliers, and accounts for feature interactions. | Does not inherently include feature selection mechanisms | $O(n \log n \times d)$ – possibly reaching a theoretical $O(n^2 \times d)$ for datasets > n = 25000 × d = 1000. |
| Stochastic Outlier Selection (SOS) | Stochastic | Considers affinity of points based on probability; Robust to outliers. | Computational complexity can be high; Parameter tuning is critical. | $O(n^2 \times d \log d)$ |
| Subspace Outlier Detection (SOD) | Subspace-based | Effective in high-dimensional data by focusing on relevant subspaces. | Can miss outliers that are not apparent in lower dimensions. | $O(n^{1.5} \times d)$ |

*Estimated via a basic examination across 100, 1000, and 10000 samples (n) with 50, 100, 500, and 1000 features (d).

## 4.0 Description of benchmarking experiments, algorithms, and datasets

A series comprising 39 synthetic and real datasets was run and evaluated in a Python 3.10.5 environment using an Intel(R) Core(TM) i7-9700F CPU @ 3.00GHz and an installed RAM of 32.0GB. To ensure reproducibility, the settings of SPINEX and the other algorithms presented earlier will be found in our Python script. All algorithms ran in default settings. The performance of all algorithms is evaluated through various performance metrics (see Table 2 [33]). These metrics include precision, recall, F1-score, and area under the receiver operating characteristic (ROC) curve (AUC).

Table 2 List of common performance metrics.

| Metric | Formula |
|---|---|
| Precision | $Precision = \dfrac{TP}{TP + FP}$ |

| Recall | $Recall = \dfrac{TP}{TP + FN}$ |
|---|---|
| F$_1$ score | $F_1 = \dfrac{2TP}{2TP + FP + FN}$ |
| Area under the ROC curve | $AUC = \displaystyle\sum_{i=1}^{N-1} \dfrac{1}{2}(FP_{i+1} - FP_i)(TP_{i+1} - TP_i)$ |

Note: P (number of real positives), N (number of real negatives), TP (true positives), TN (true negatives), FP (false positives), and FN (false negatives).

Each metric provides unique insights from identifying anomalies (often regarded as the minority class). For example, precision measures the accuracy of the positive predictions made by the algorithm and quantifies explicitly the proportion of data points that were correctly identified as anomalies (true positives) out of all the data points that were predicted as anomalies (true positives and false positives). High precision indicates that the algorithm correctly made most of its "anomaly" predictions, minimizing false alarms. On the other hand, recall is defined as the ratio of true positives to the sum of true positives and false negatives (anomalies that the algorithm failed to detect). This metric measures the algorithm's ability to detect all actual anomalies.

The F1-score is the harmonic mean of precision and recall and hence combines both precision and recall into a single measure to capture the balance between them. Since it accounts for both false positives and false negatives, this metric becomes useful for understanding an algorithm's overall accuracy without favoring either precision or recall. The AUC ROC illustrates the algorithm for discriminating between the classes across all possible threshold values. The ROC curve is a plot of the true positive rate (recall) against the false positive rate (the ratio of incorrectly labeled normal instances to all actual normal instances) for various threshold settings. AUC values range from 0 to 1, with 1 indicating perfect classification and 0.5 denoting no discriminative ability.

*4.1 Synthetic datasets*

Twenty one synthetic datasets were used in our experiments. Each dataset simulates different conditions and scenarios for anomaly detection purposes. These datasets were created using the following specially designed function (see Fig. 2 and Table 3).

generate_complex_data: This function is designed to simulate data with varying degrees of complexity, incorporating both normal data points and outliers. Key parameters include mean_shift, cov_scale, outlier_fraction, size, num_features, and complexity_level, which allow for extensive control over the data characteristics:

- mean_shift: Shifts the mean of the normal data distribution
- cov_scale: Scales the covariance matrix for both normal data and outliers
- outlier_fraction: Determines the proportion of outliers in the dataset
- size: Total number of data points to generate
- num_features: Number of features in the base dataset
- complexity_level: Controls the addition of nonlinear interactions and higher-order terms

The function begins by setting a random seed for reproducibility. It then initializes the base mean vector and covariance matrix for the normal data distribution. The base mean is a zero vector, and the base covariance is an identity matrix scaled by cov_scale. The number of outliers is calculated based on the outlier_fraction and size parameters. The function uses math.ceil() to generate at least one outlier when the calculated number is non-zero but less than one.

The function then introduces additional complexity based on the complexity_level parameter:
- For complexity_level > 0, nonlinear interactions are added:
  - The product of the first two features is calculated and appended as a new feature: $f_{new} = f_1 \times f_2$
- For complexity_level > 1, higher-order terms are added via:
  - Squared terms of all features: $f_{new} = f_1^2$
  - Sine transformation of all features: $f_{new} = sin(f_i)$

After generating the data, the function combines normal data and outliers and creates corresponding labels (0 for normal data, 1 for outliers). Finally, it shuffles the data and labels to ensure random ordering.

Table 3 Parameters used in the synthetic datasets

| No. | Mean Shift | Covariance Scale | Outlier Fraction | Number of Features | Complexity Level | Size |
| --- | --- | --- | --- | --- | --- | --- |
| 1 | 4 | 1.2 | 0.03 | 3 | 0 | 100 |
| 2 | -1 | 0.7 | 0.04 | 9 | 1 | 350 |
| 3 | 5 | 1.1 | 0.12 | 5 | 2 | 8000 |
| 4 | -4 | 0.6 | 0.08 | 3 | 0 | 550 |
| 5 | 2 | 2.5 | 0.07 | 11 | 0 | 120 |
| 6 | -3 | 0.3 | 0.18 | 19 | 1 | 300 |
| 7 | 1.5 | 0.9 | 0.11 | 16 | 1 | 400 |
| 8 | -2 | 1.3 | 0.16 | 10 | 2 | 100 |
| 9 | -1 | 0.4 | 0.19 | 4 | 2 | 320 |
| 10 | 4.5 | 0.9 | 0.06 | 25 | 1 | 4200 |
| 11 | -4.5 | 0.7 | 0.09 | 11 | 0 | 520 |
| 12 | 2.5 | 1.6 | 0.13 | 14 | 0 | 130 |
| 13 | -3.5 | 0.5 | 0.11 | 10 | 2 | 590 |
| 14 | 1.2 | 1.7 | 0.17 | 18 | 1 | 1400 |
| 15 | 3.5 | 1.2 | 0.15 | 13 | 1 | 440 |
| 16 | -1.5 | 0.8 | 0.22 | 150 | 0 | 5040 |
| 17 | 2 | 1 | 0.03 | 3 | 0 | 200 |
| 18 | -2 | 0.5 | 0.04 | 13 | 1 | 3000 |
| 19 | -1 | 0.7 | 0.05 | 9 | 2 | 150 |
| 20 | 3 | 0.6 | 0.02 | 7 | 1 | 250 |
| 21 | -3 | 1.1 | 0.06 | 30 | 0 | 1000 |

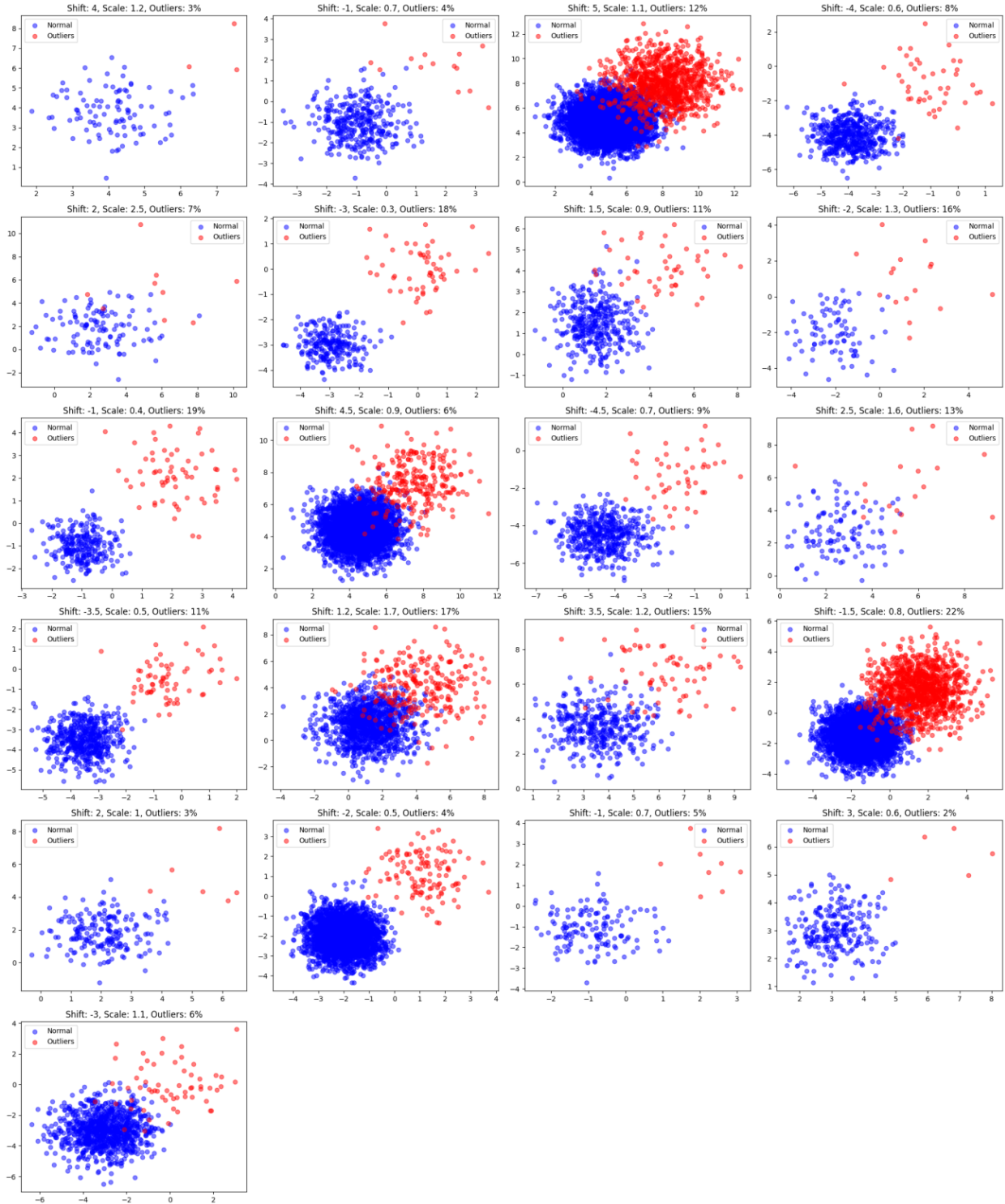

Fig. 2 Visualization of the synthetic datasets

We followed a systematic methodology to rank the examined algorithms' performance across all listed datasets. This methodology calculates the average scores for each algorithm on each dataset and across all metrics. Then, we rank algorithms per metric, calculate the sum of ranks across all metrics, and display the ranked models. The outcome of this analysis is shown in Table 4 as well as Fig. 3. It is quite clear that SPINEX and most of its variants rank well when compared to other algorithms. Compared with other well-known algorithms like HBOS, LOF, and various k-NN based methods, most SPINEX variants generally exhibit competitive or superior rankings (with nuanced differences appearing between the variants that incorporate weights and interactions vs. those of dynamic thresholds). This competitive edge likely stems from SPINEX's unique approach to anomaly detection, which might include inherent processing and interaction utilization methods.

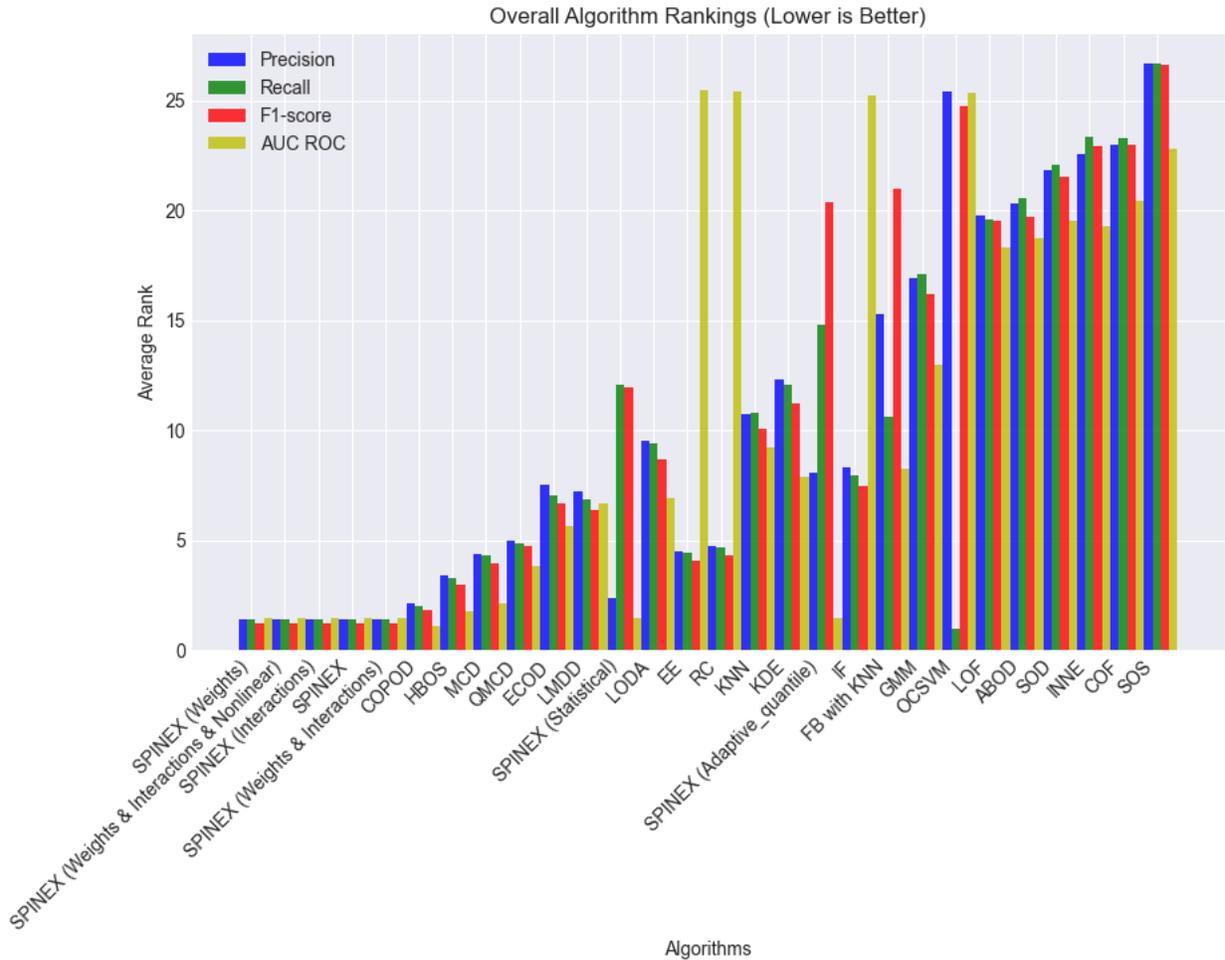

Fig. 3 Outcome of benchmarking across synthetic datasets

Table 4 Average and overall ranking results on synthetic data

| Algorithm | Precision | Recall | F1-score | AUC ROC | Overall |
|---|---|---|---|---|---|
| SPINEX | 1.429 | 1.381 | 1.238 | 1.476 | 1 |
| COPOD | 2.143 | 2.000 | 1.810 | 1.095 | 2 |
| HBOS | 3.381 | 3.286 | 2.952 | 1.762 | 3 |
| MCD | 4.381 | 4.333 | 3.952 | 2.143 | 4 |
| QMCD | 4.952 | 4.857 | 4.714 | 3.810 | 5 |
| ECOD | 7.524 | 7.048 | 6.667 | 5.619 | 6 |
| LMDD | 7.190 | 6.857 | 6.381 | 6.667 | 7 |
| SPINEX (Statistical) | 2.381 | 12.095 | 11.952 | 1.476 | 8 |
| LODA | 9.524 | 9.381 | 8.667 | 6.905 | 9 |
| EE | 4.476 | 4.429 | 4.048 | 25.476 | 10 |
| RC | 4.714 | 4.667 | 4.333 | 25.429 | 11 |
| KNN | 10.762 | 10.810 | 10.095 | 9.190 | 12 |
| KDE | 12.333 | 12.095 | 11.238 | 7.857 | 13 |
| SPINEX (Adaptive_quantile) | 8.048 | 14.810 | 20.381 | 1.476 | 14 |
| IF | 8.286 | 7.952 | 7.476 | 25.190 | 15 |
| FB with KNN | 15.286 | 10.619 | 21.000 | 8.238 | 16 |
| GMM | 16.905 | 17.095 | 16.190 | 13.000 | 17 |
| OCSVM | 25.429 | 1.000 | 24.762 | 25.333 | 18 |
| LOF | 19.762 | 19.571 | 19.524 | 18.286 | 19 |
| ABOD | 20.333 | 20.524 | 19.714 | 18.762 | 20 |
| SOD | 21.810 | 22.095 | 21.524 | 19.524 | 21 |
| INNE | 22.571 | 23.333 | 22.905 | 19.286 | 22 |
| COF | 23.000 | 23.286 | 23.000 | 20.429 | 23 |
| SOS | 26.667 | 26.667 | 26.619 | 22.810 | 24 |

Figure 4 utilizes the Principal Component Analysis (PCA) to reduce the data to two dimensions as a means to allow us to visually assess how each algorithm separates normal points from outliers. This plot shows such a plot for two datasets (no. 1 and no. 4). The standard SPINEX shows a clear separation between normal data and outliers, with the boundary being relatively close to the cluster of normal points. This suggests a careful approach to anomaly detection, as that can also be seen in other algorithms such as IF, KNN, GMM, EE etc. On the other hand, other algorithms, such as ABOD, SOD, COPOD, INNE, etc., seem to struggle to separate the outliers. Similar plots for all datasets are shown in the Appendix.

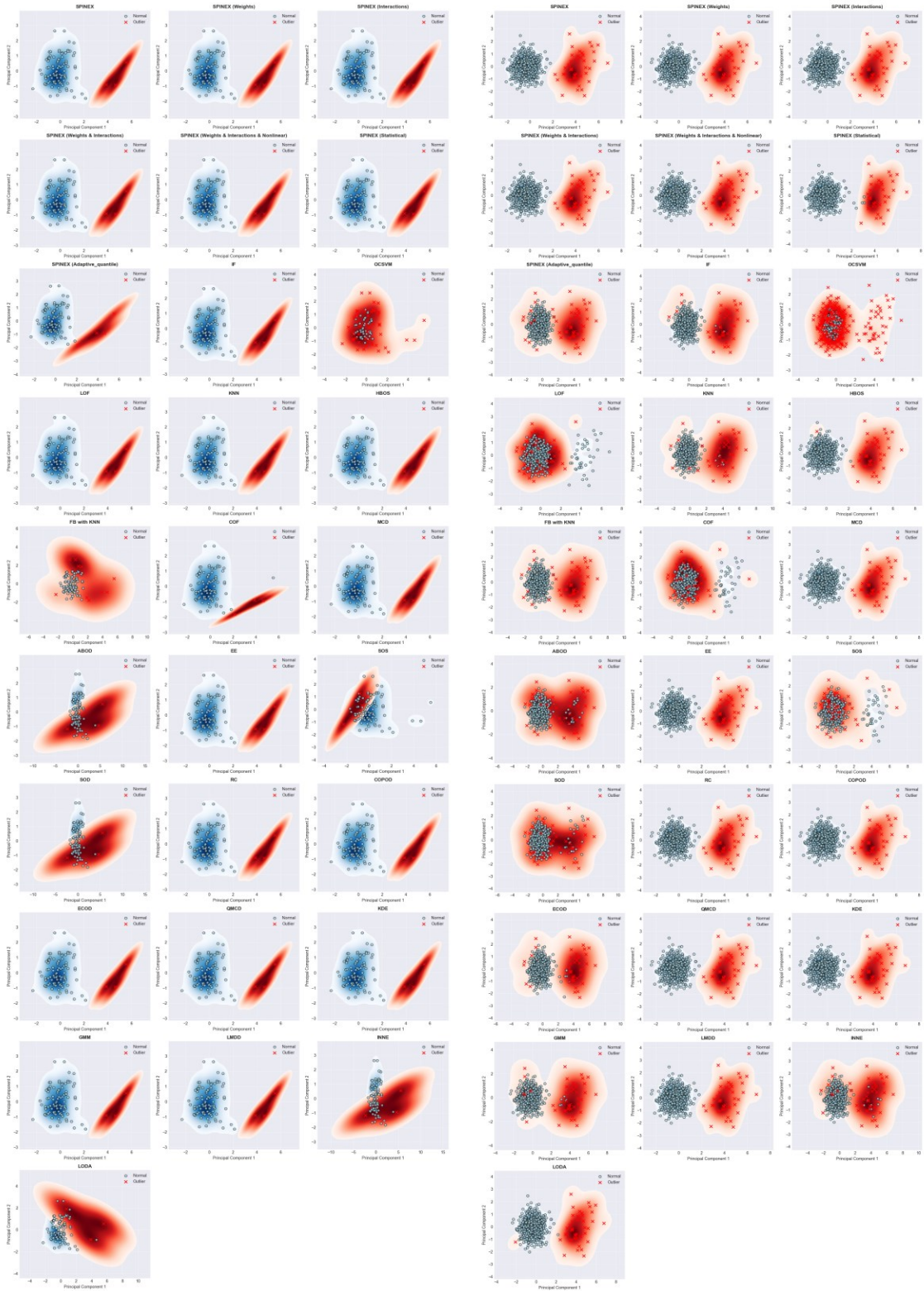

Fig. 4 Visualization of anomalies predicted across two dataset scenarios (no. 1 [left] and no. 4 [right])

## 4.2 Real datasets

This section offers insights into the same systemic analysis followed on the synthetic datasets but by using 18 real datasets. These datasets consist of various problems and scenarios and are described in Table 5. This table lists details on each dataset, along with their respective references. Additional details can be found in the respective references. Notably, many of these datasets were recommended by the following notable and comprehensive benchmarking studies [34–36] – see Figure 5.

Table 5 Real datasets used in the analysis

| Dataset | No. of samples | No. of features | % Anomaly | Domain | Ref. |
|---|---|---|---|---|---|
| ALOI | 49,534 | 27 | 3.04 | Image | [35] |
| Annthyroid | 7200 | 6 | 7.42 | Healthcare | [35,37] |
| Arrhythmia | 450 | 259 | 2.00 | Healthcare | [35] |
| Bank | 41,188 | 10 | 11.2 | Finance | [36] |
| Cardio | 1831 | 21 | 9.61 | Healthcare | [35,38] |
| Glass | 214 | 7 | 4.21 | Forensic | [35] |
| HeartDisease | 270 | 13 | 2.00 | Healthcare | [35] |
| Hepatitis | 80 | 19 | 16.25 | Healthcare | [35] |
| Ionosphere | 351 | 33 | 35.90 | Oryctognosy | [35,39] |
| PageBlocks | 5393 | 10 | 9.46 | Document | [35] |
| Parkinson | 195 | 22 | 2.00 | Healthcare | [35] |
| Pendigits | 6870 | 16 | 2.27 | Image | [35,40] |
| Pima | 768 | 8 | 34.90 | Healthcare | [35] |
| SpamBase | 4207 | 57 | 39.91 | Document | [35] |
| Stamps | 340 | 9 | 9.12 | Document | [35] |
| Waveform | 3443 | 21 | 2.90 | Physics | [35,41] |
| Wilt | 4819 | 5 | 5.33 | Botany | [35] |
| WPBC | 198 | 33 | 23.74 | Healthcare | [35,42] |

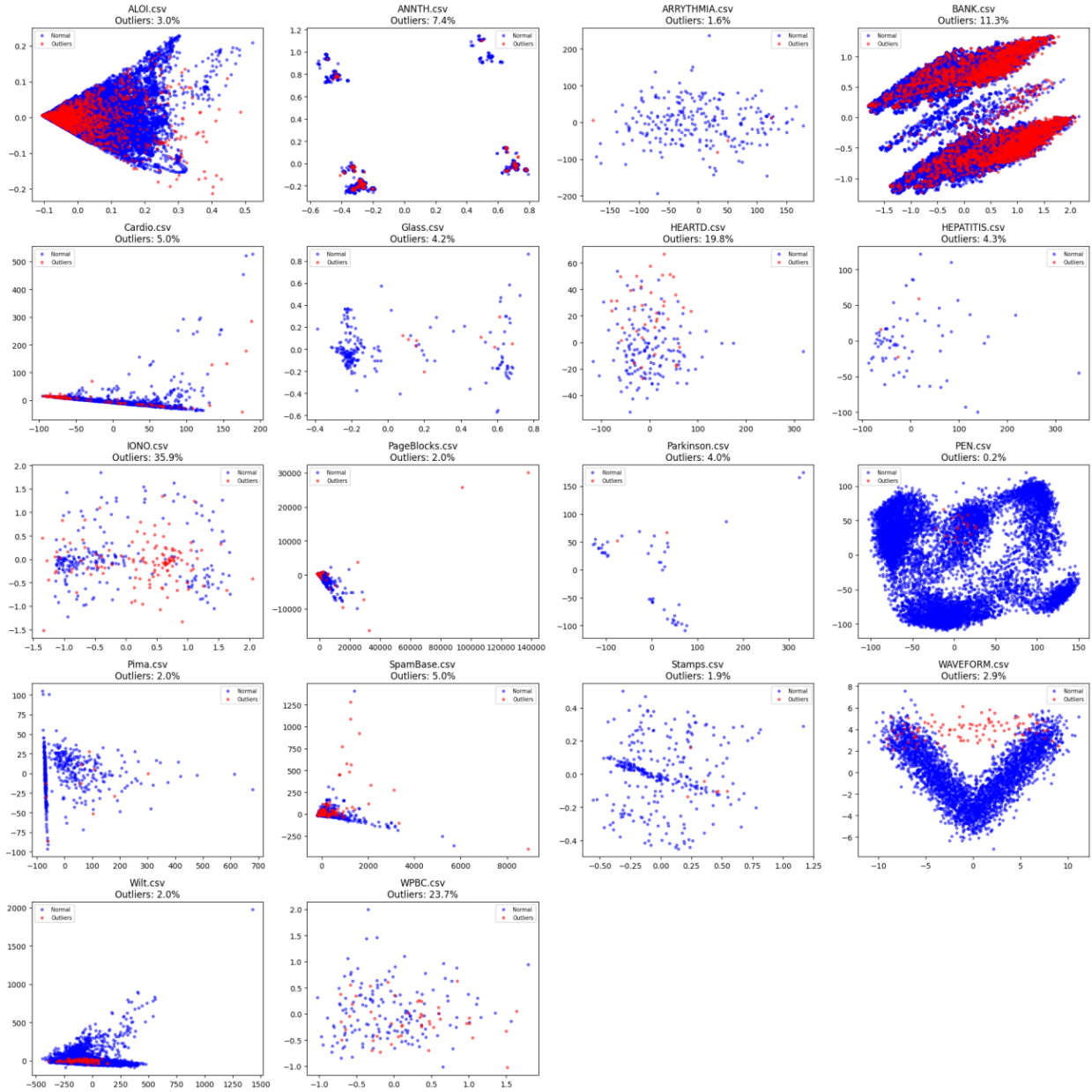

Fig. 5 Visualization of the real datasets

The results of the benchmarking and ranking analysis are listed in Table 6 and shown in Fig. 6. These results clearly show that the SPINEX algorithm and its variants perform comparatively well against the other commonly used anomaly detection algorithms. For example, these variants rank within the top-10 performing algorithms. The same results also note that the use of additional settings (in terms of weights and interactions) does not seem to offer significant variations in the average rankings. It is quite clear that the KNN-like algorithms, along with SOD, COPOD, INNE, and KDE, rank on the top-5 list of algorithms.

374    Table 6 Average and overall ranking results on real data

| Algorithm | Precision | Recall | F1-score | AUC ROC | Overall |
|---|---|---|---|---|---|
| KNN | 8.222 | 8.556 | 7.944 | 6.167 | 1 |
| SOD | 8.389 | 9.056 | 8.111 | 9.389 | 2 |
| COPOD | 9.944 | 10.333 | 9.722 | 6.278 | 3 |
| FB with KNN | 12.611 | 7.611 | 11.833 | 6.111 | 4 |
| INNE | 9.556 | 9.889 | 9.111 | 9.722 | 5 |
| KDE | 9.889 | 10.278 | 9.667 | 9.167 | 6 |
| SPINEX | 8.667 | 8.833 | 8.167 | 14.000 | 7 |
| GMM | 10.167 | 10.222 | 9.778 | 11.556 | 8 |
| MCD | 11.056 | 11.111 | 10.611 | 9.500 | 9 |
| HBOS | 11.000 | 11.444 | 11.000 | 9.500 | 10 |
| COF | 11.000 | 11.444 | 10.778 | 10.444 | 11 |
| LMDD | 10.000 | 10.500 | 9.889 | 13.944 | 12 |
| ECOD | 12.833 | 12.833 | 12.333 | 9.667 | 13 |
| QMCD | 12.556 | 12.944 | 12.222 | 10.722 | 14 |
| LOF | 10.556 | 11.000 | 10.278 | 18.667 | 15 |
| SPINEX (Adaptive_quantile) | 8.833 | 12.667 | 15.944 | 14.000 | 16 |
| EE | 9.167 | 9.333 | 8.667 | 24.833 | 17 |
| IF | 9.389 | 9.889 | 9.111 | 24.500 | 18 |
| ABOD | 13.824 | 14.000 | 13.412 | 11.765 | 19 |
| SPINEX (Statistical) | 10.500 | 13.556 | 16.611 | 14.000 | 20 |
| LODA | 14.167 | 14.222 | 14.000 | 15.556 | 21 |
| OCSVM | 18.833 | 1.444 | 14.778 | 23.722 | 22 |
| RC | 12.444 | 12.333 | 11.944 | 24.611 | 23 |
| SOS | 15.333 | 15.444 | 15.056 | 16.833 | 24 |

375

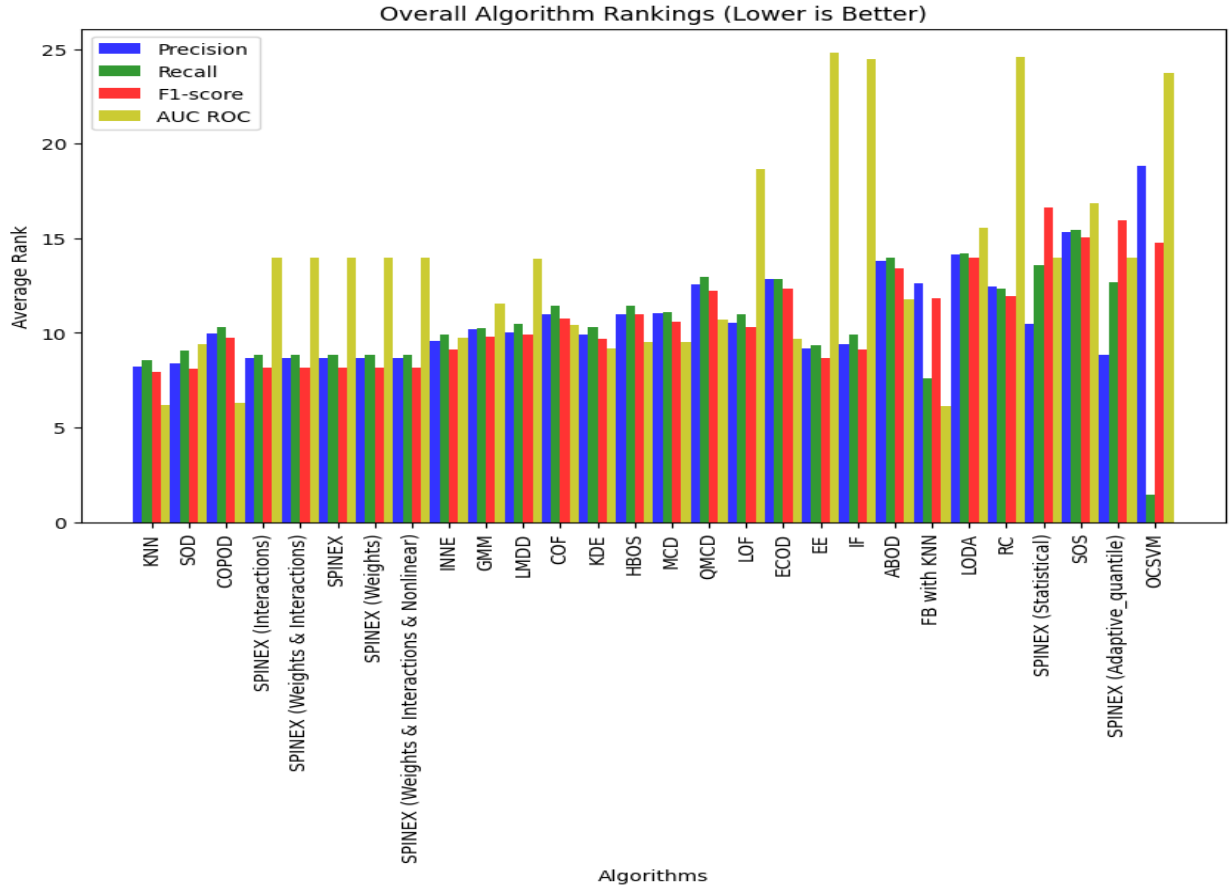

Fig. 6 Rankings on real datasets

Figure 7 presents two visual examples by displaying the PCA for the Pendigits and Waveform datasets. This plot shows that SPINEX clearly identifies outliers, unlike other algorithms, such as OCSVM and FB with KNN. These algorithms seem to struggle to separate the outliers. Please note that the plots for all datasets are shown in the Appendix.

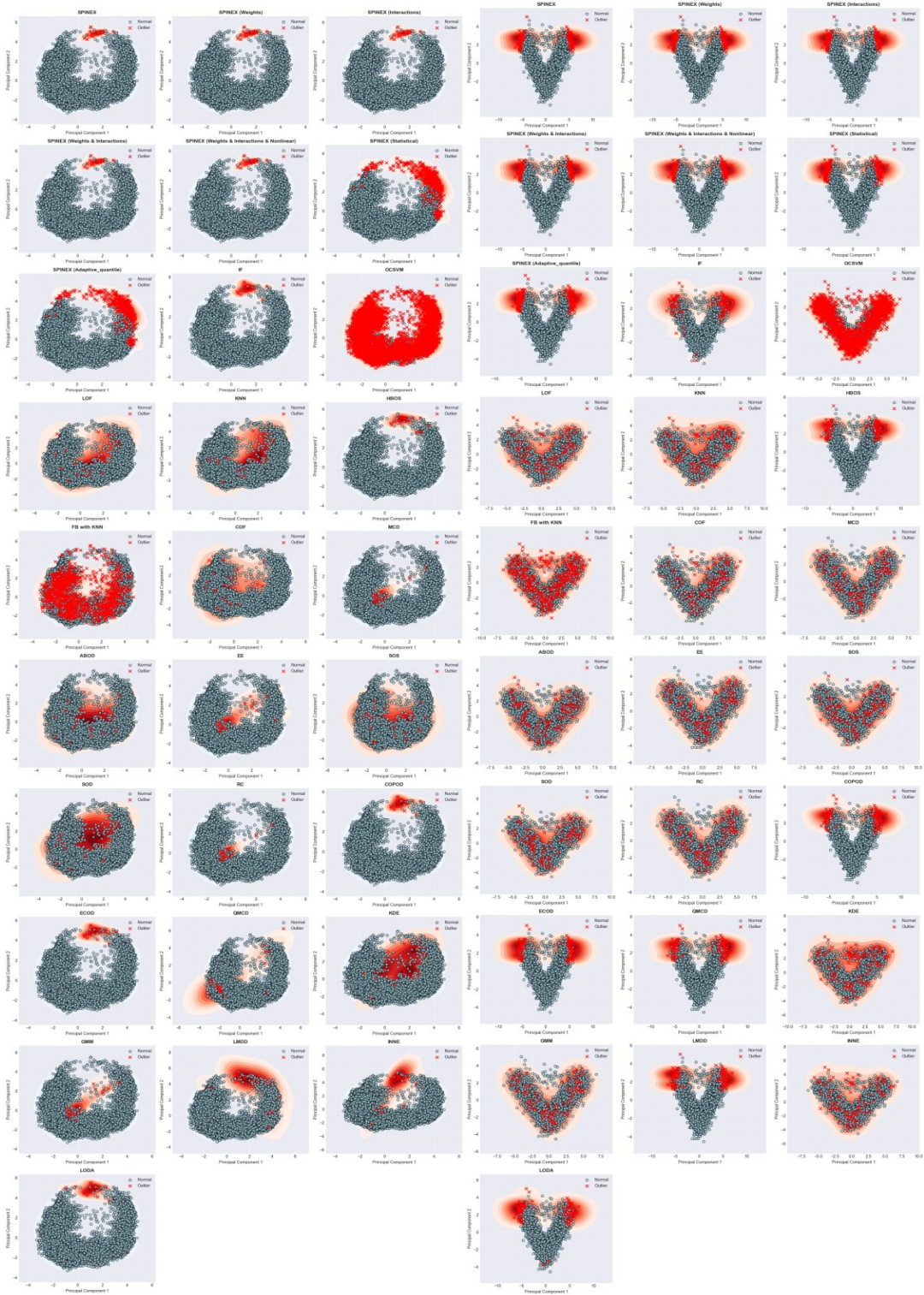

Fig. 7 Visualization of anomalies predicted across two real dataset scenarios (no. 1 [Pendigits] and no. 4 [Waveform])

## 5.0 A note on explainability and future research directions

To showcase the explainability performance of SPINEX, two anomalous data points (i.e., No. 43 and no. 93) of the three identified from the first synthetic dataset are examined herein. Figure 8 shows the calculated feature importance and contribution (by using calculate_feature_contributions method) to each anomalous data point. As one can see, these data points yielded significant values that were above the baseline established for anomaly scores. This visualization can help guide users on how a given data point was predicted to be anomalous or normal.

**Anomaly at index 43:**
  - Feature3: 8.39 (baseline: 4.18, contribution: 4.21)
  - Feature1: 7.55 (baseline: 4.19, contribution: 3.36)
  - Feature2: 5.93 (baseline: 3.89, contribution: 2.04)

**Anomaly at index 93:**
  - Feature2: 8.26 (baseline: 3.89, contribution: 4.37)
  - Feature3: 7.98 (baseline: 4.18, contribution: 3.79)
  - Feature1: 7.48 (baseline: 4.19, contribution: 3.28)

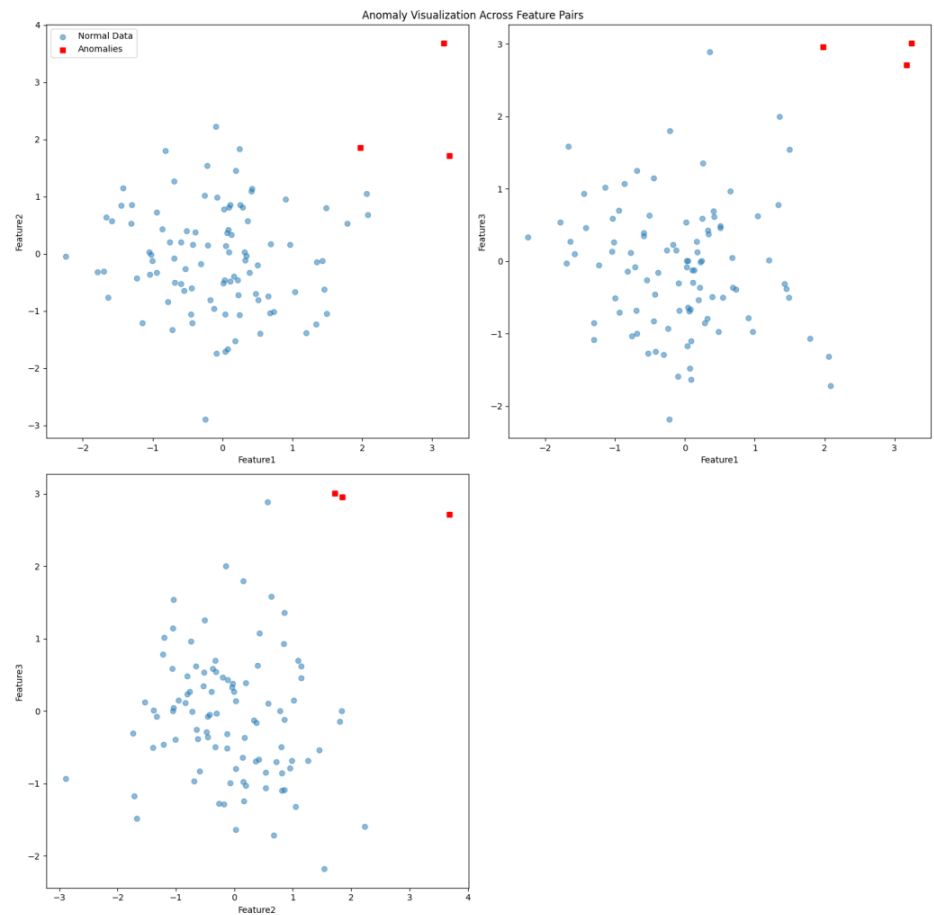

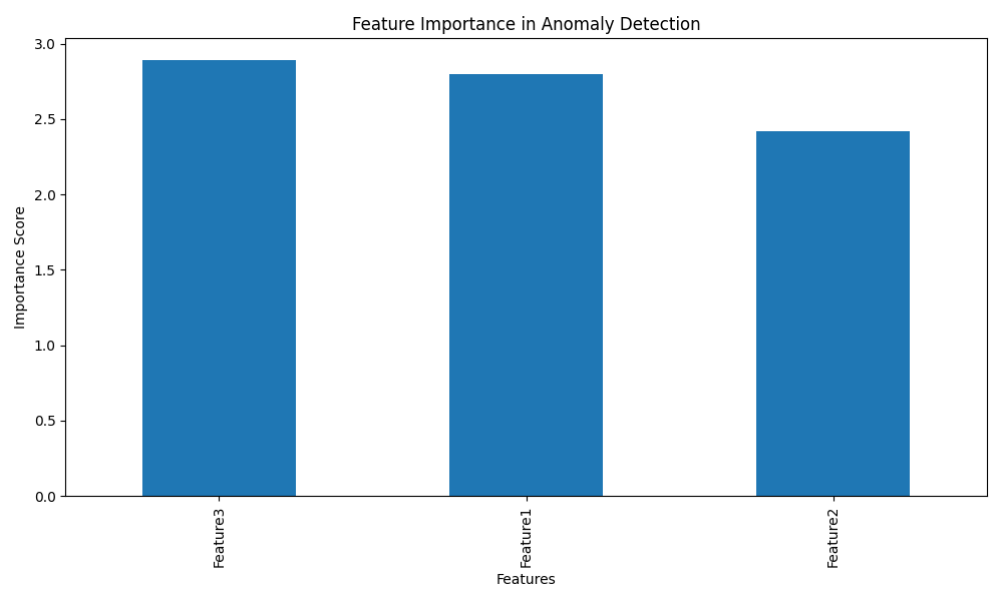

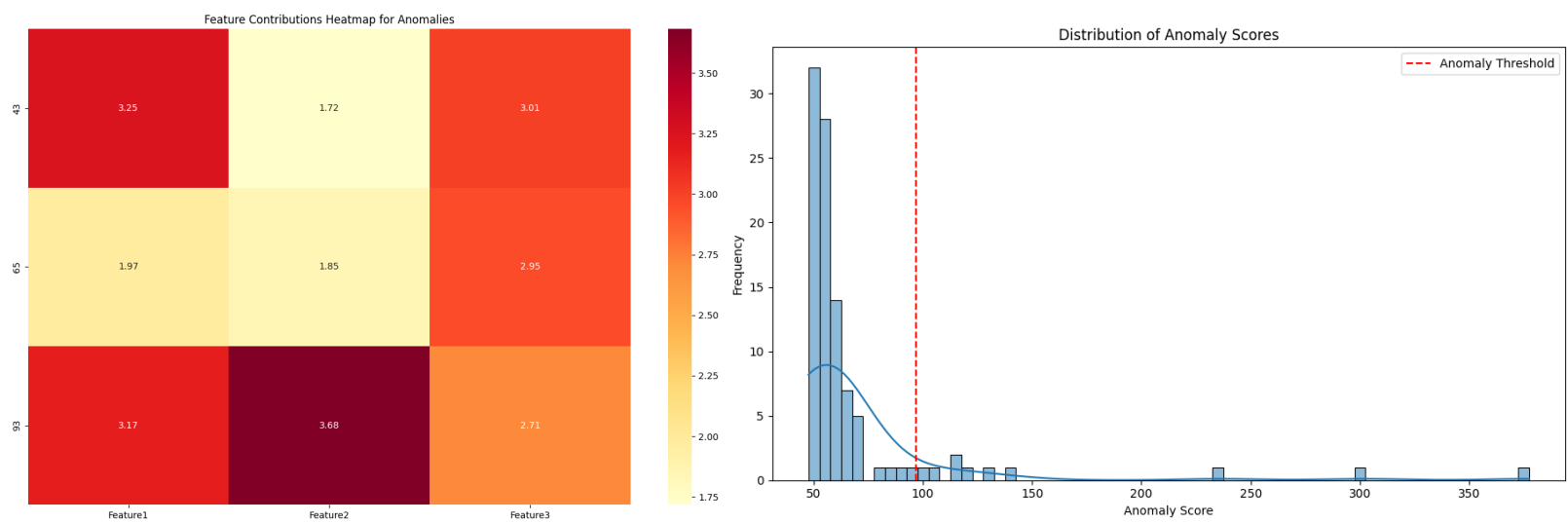

Fig. 8 Example of explainability

Despite the significant progress made in anomaly detection over the past decades, several challenges persist, limiting the effectiveness of existing methods and motivating continued research in this field. Such challenges apply not only to SPINEX but to many of the examined algorithms as well. Hence, understanding these challenges is crucial for future improvements [43].

One of the primary challenges lies in the inherent imbalance in datasets. Anomalies, by definition, are rare occurrences, and hence, they comprise only a tiny fraction of the data. This imbalance can lead to difficulties in training models, especially supervised ones, and may result in biased or overfitted models. Another challenge is the dynamic nature of normal behavior in real-world scenarios. What constitutes normal can evolve over time (a phenomenon known as concept drift). In some scenarios, this evolution can cause previously developed algorithms to become ineffective or outdated. Thus, developing algorithms that can adapt to these changes while maintaining high detection accuracy remains an open problem in the field [44].

The high-dimensional nature of many modern datasets poses another set of challenges. As the number of features increases, the sparsity of data in this high-dimensional space makes it difficult to distinguish between normal points and anomalies (a phenomenon known as the curse of dimensionality). Moreover, in high-dimensional spaces, the notion of similarity may become less intuitive [45]. To handle such high-dimensional data, scalability is an increasingly important concern as the volume and velocity of data continue to grow. Thus, developing algorithms that can handle big data while maintaining high accuracy and low latency is a critical area of ongoing research.

The interpretability of anomaly detection results is another pressing challenge, particularly in high-stakes applications (e.g., healthcare or financial fraud detection). While some algorithms offer clear interpretability, there is a need for the adoption of advanced interpretability and explainability techniques. We invite interested readers to spearhead efforts aimed at overcoming such challenges.

**6.0 Conclusions**

This study introduces a new algorithm from the SPINEX (Similarity-based Predictions with Explainable Neighbors Exploration) family, designed for anomaly and outlier detection. It employs similarity metrics across multiple subspaces to effectively pinpoint outliers. The performance of SPINEX was rigorously assessed through a series of experiments, comparing it against 21 well-established anomaly detection algorithms, including Angle-Based Outlier Detection (ABOD), Connectivity-Based Outlier Factor (COF), Copula-Based Outlier Detection (COPOD), ECOD, Elliptic Envelope (EE), Feature Bagging with KNN, Gaussian Mixture Models (GMM), Histogram-based Outlier Score (HBOS), Isolation Forest (IF), Isolation Neural Network Ensemble (INNE), Kernel Density Estimation (KDE), K-Nearest Neighbors (KNN), Lightweight Online Detector of Anomalies (LODA), Linear Model Deviation-based Detector (LMDD), Local Outlier Factor (LOF), Minimum Covariance Determinant (MCD), One-Class SVM (OCSVM), Quadratic MCD (QMCD), Robust Covariance (RC), Stochastic Outlier Selection (SOS), and Subspace Outlier Detection (SOD). These comparisons were conducted across 39 benchmark datasets, both synthetic and real, covering various domains and featuring diverse dimensions and complexities.

The findings confirm that SPINEX consistently outperforms conventional anomaly detection algorithms, ranking among the top seven in effectiveness.

**Data Availability**

Some or all data, models, or code that support the findings of this study are available from the corresponding author upon reasonable request.

SPINEX can be accessed from [**to be added**].

**Conflict of Interest**

The authors declare no conflict of interest.

Appendix

Python script, additional results, and visualizations.

A. Python script: to be provided.

B. Visualization of synthetic datasets

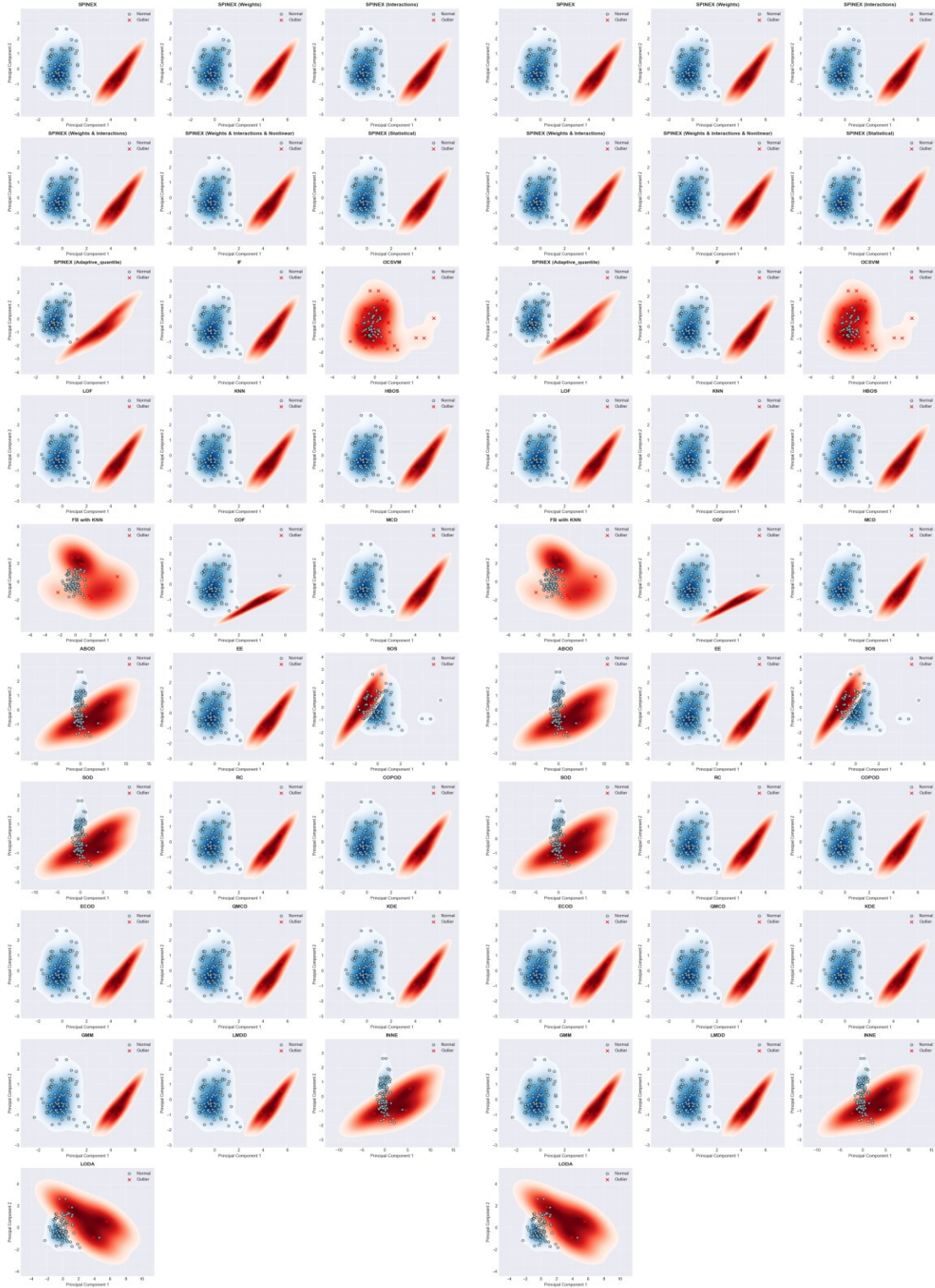

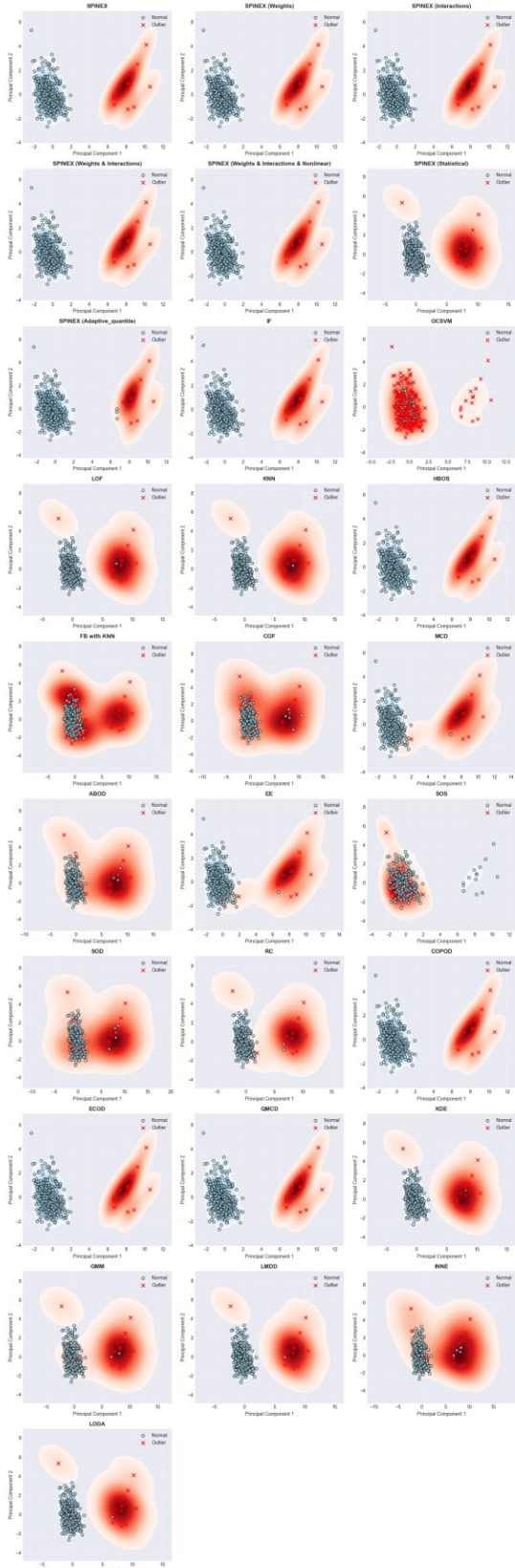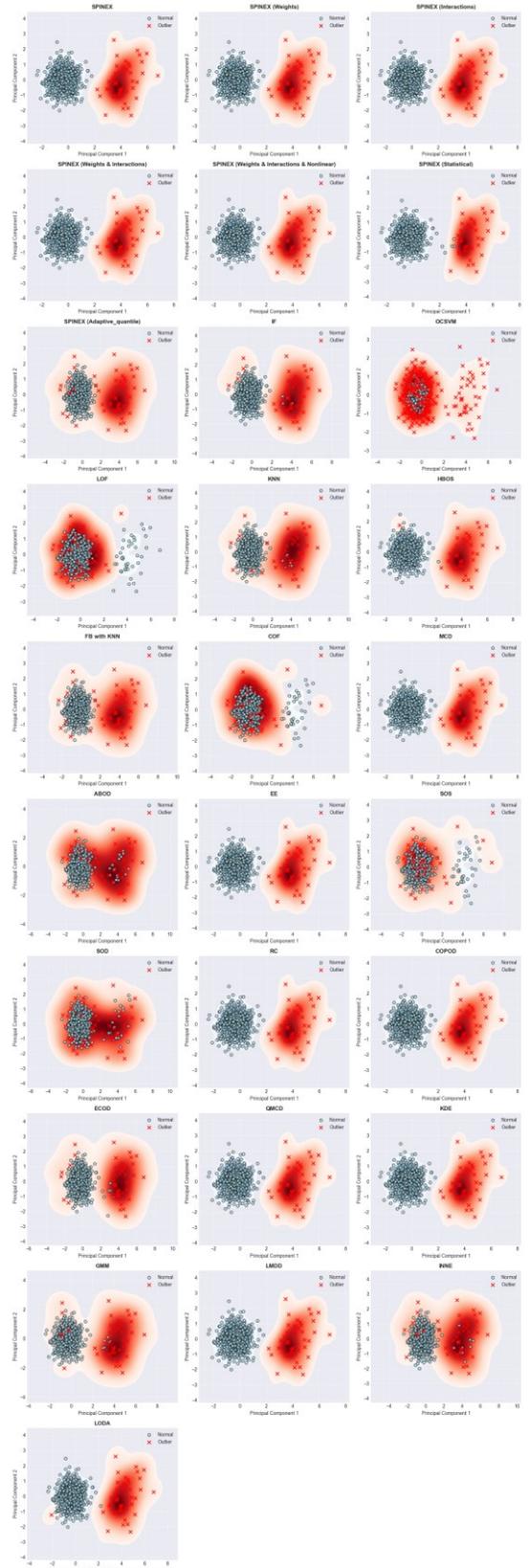

592    A.

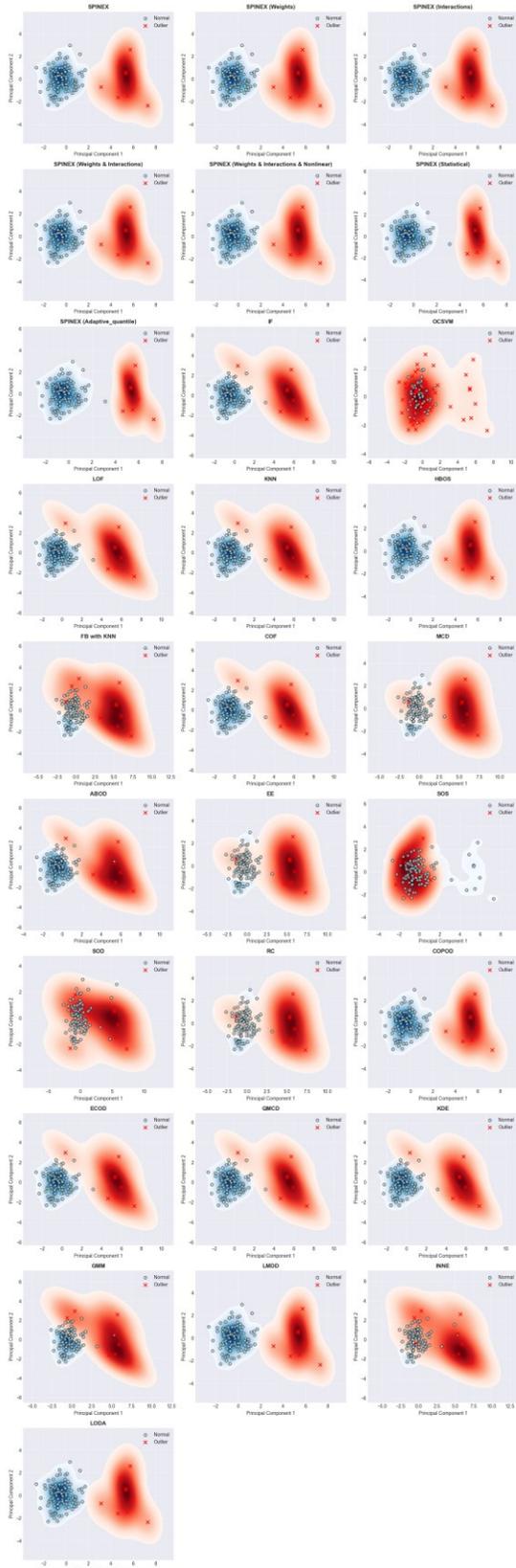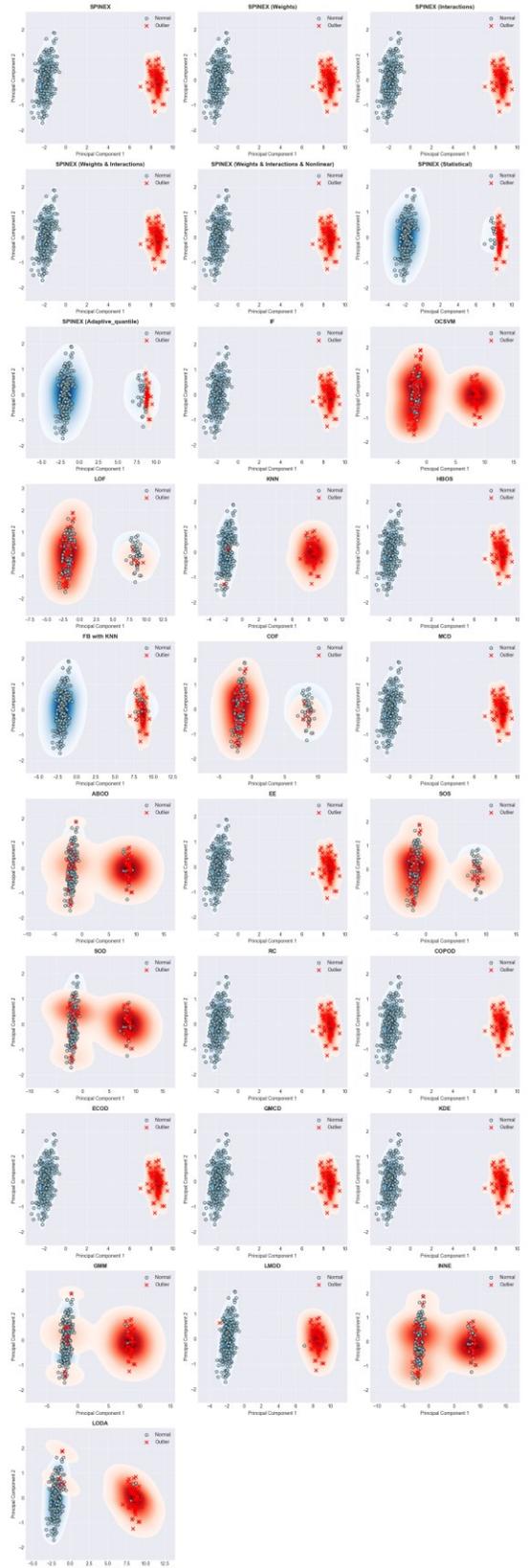

593

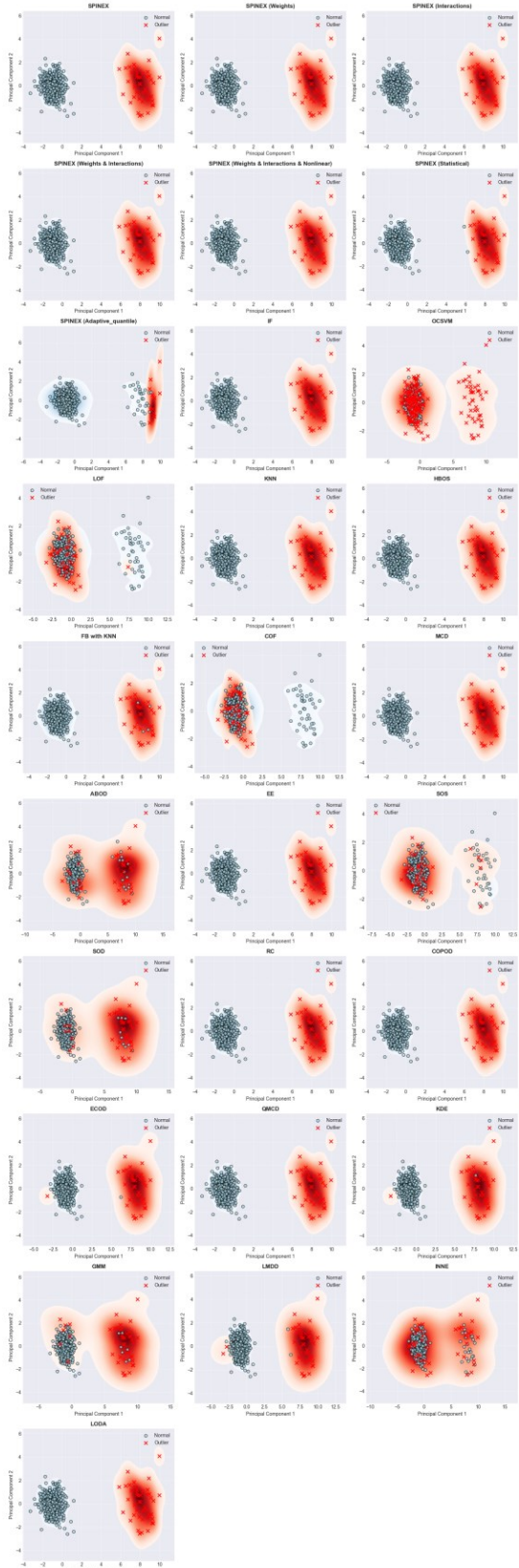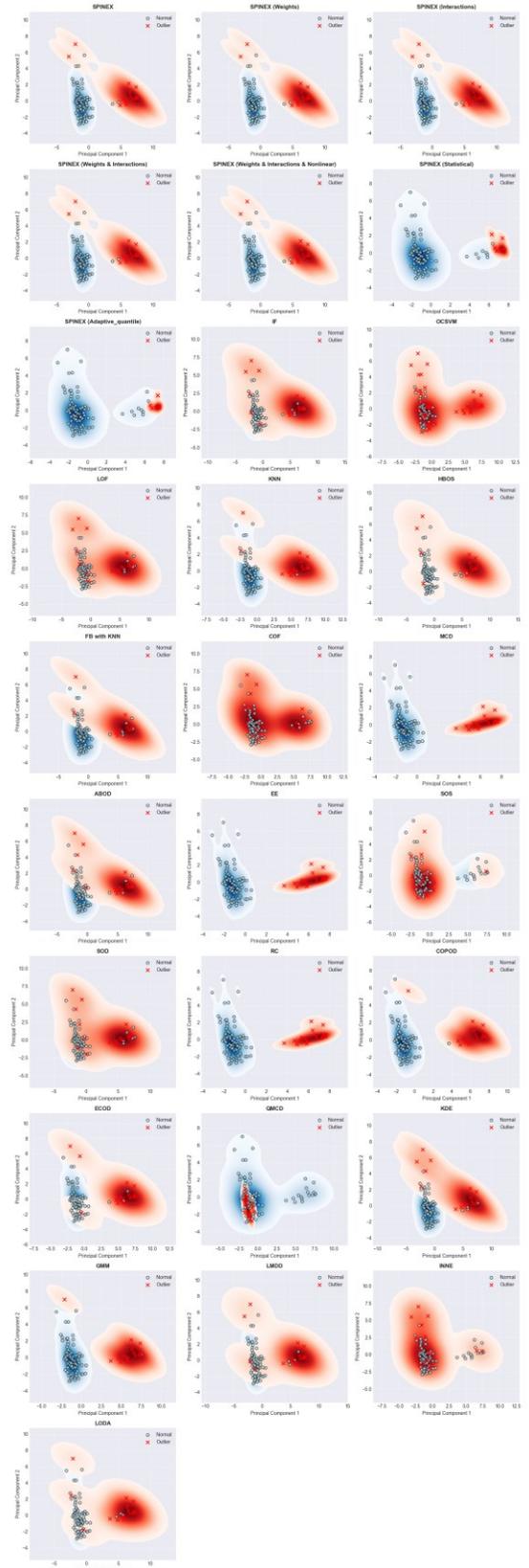

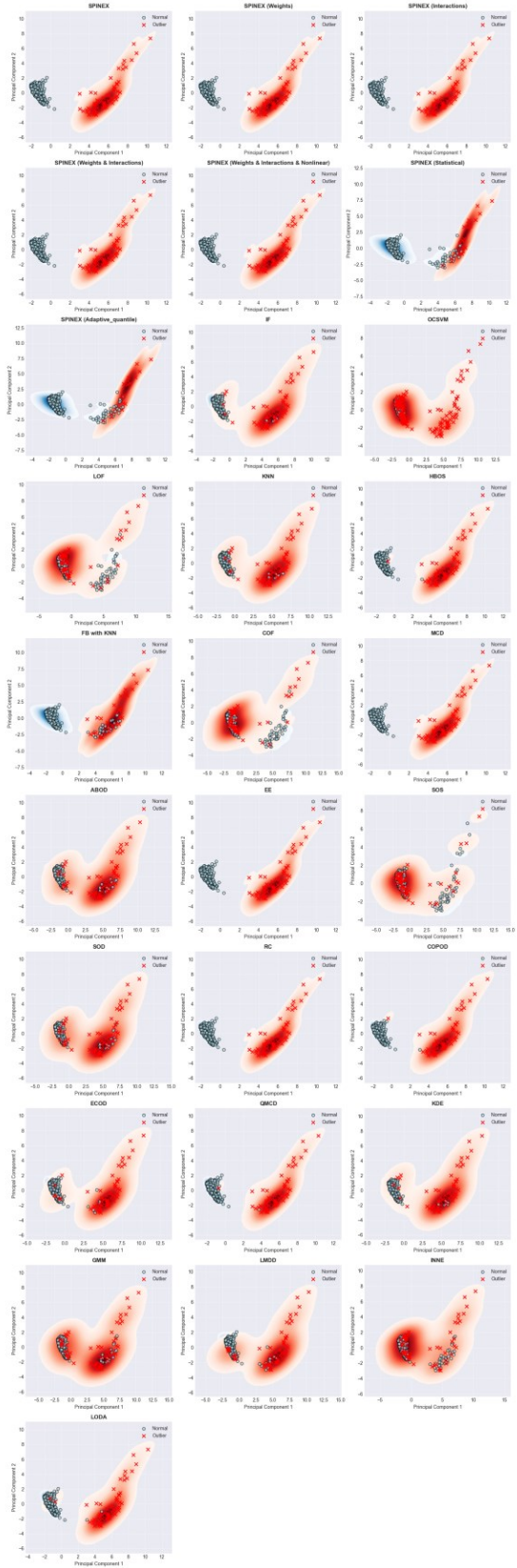
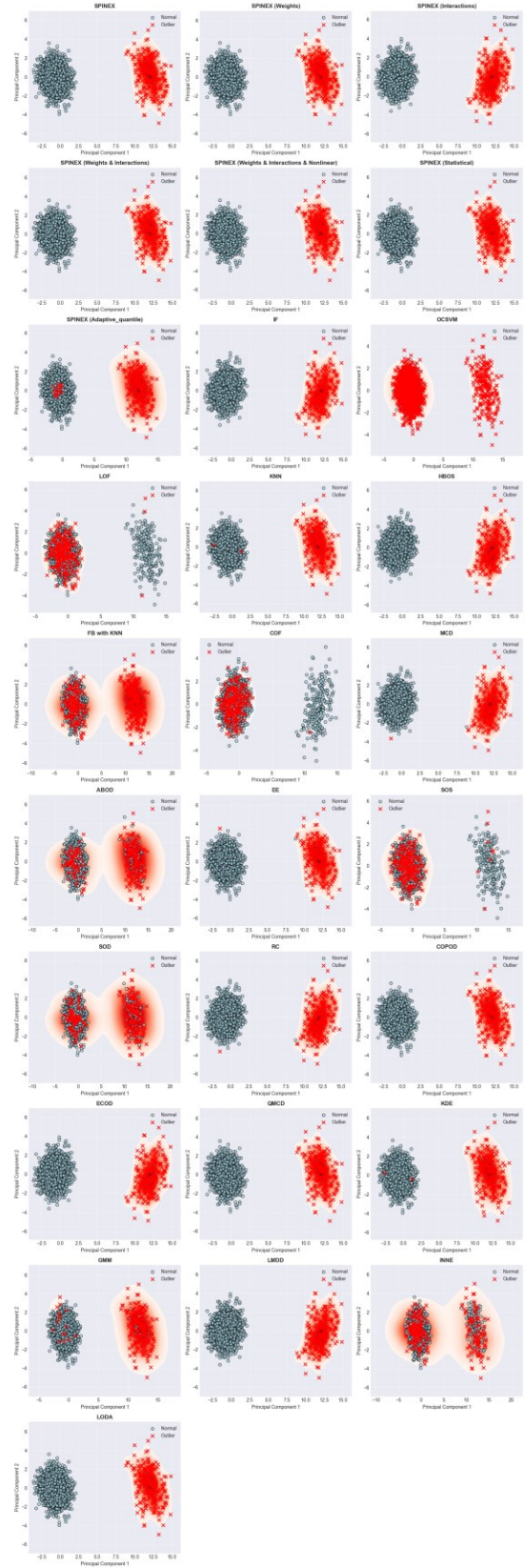

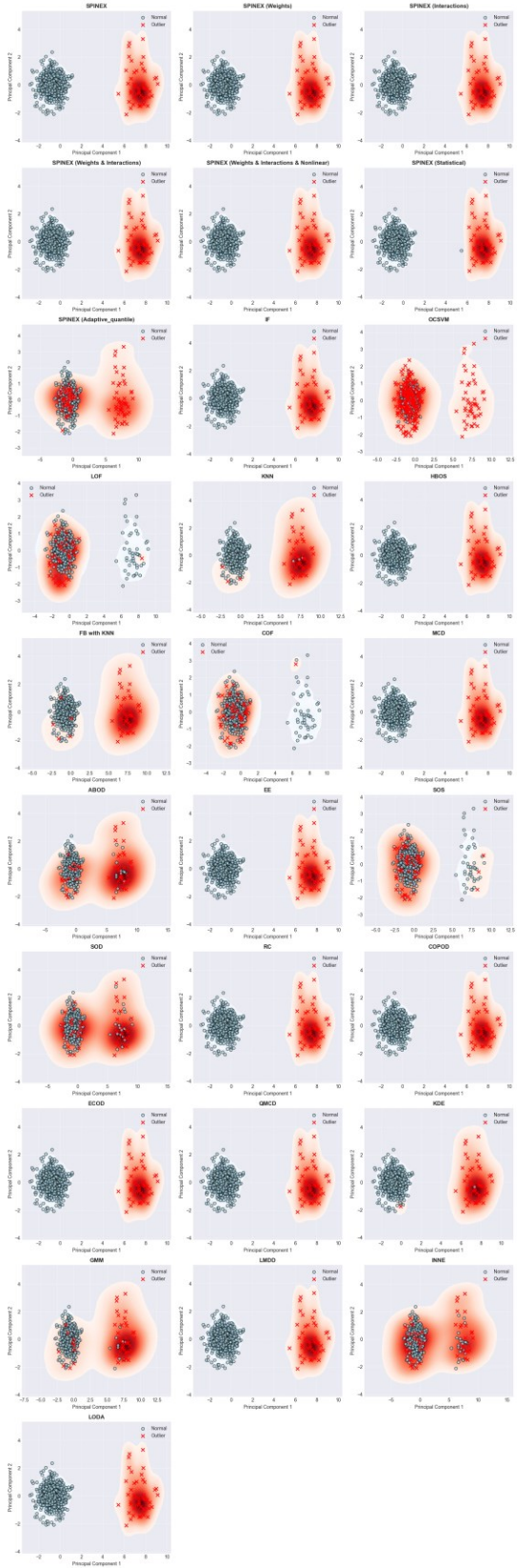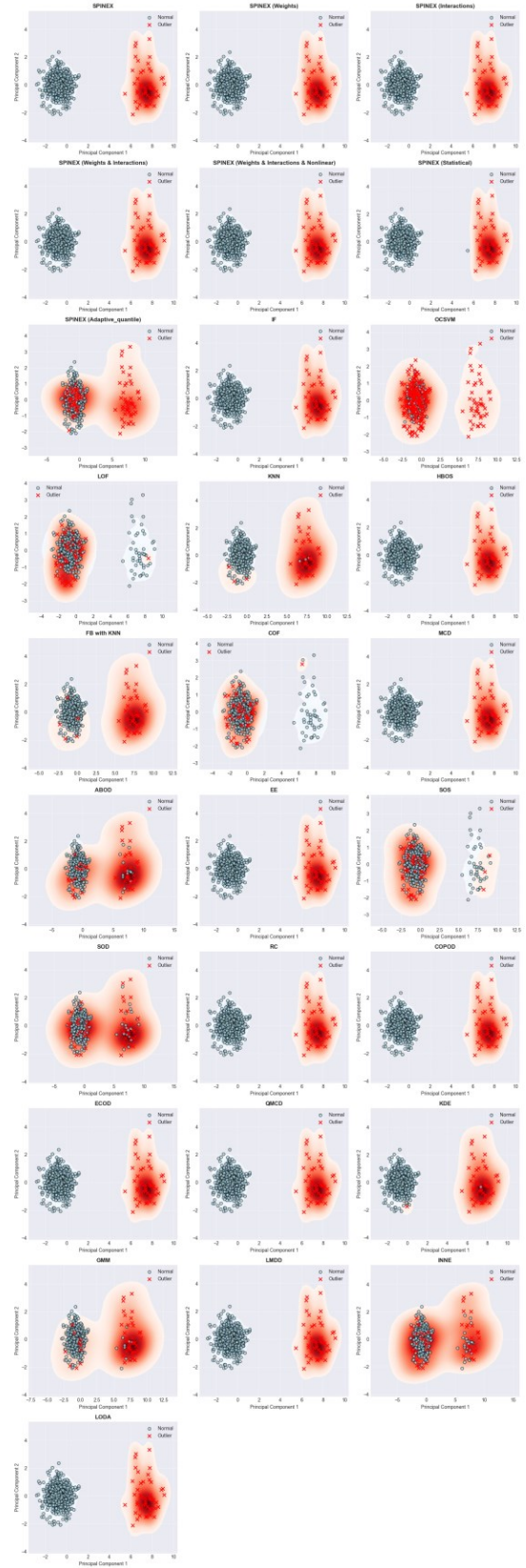

596

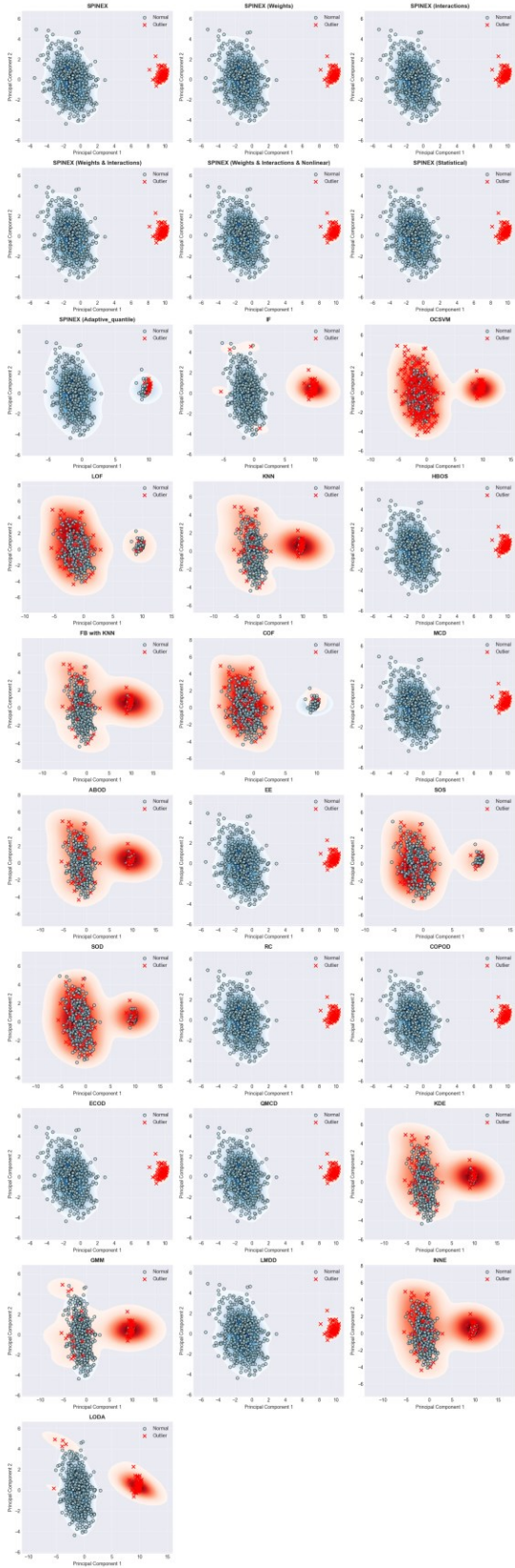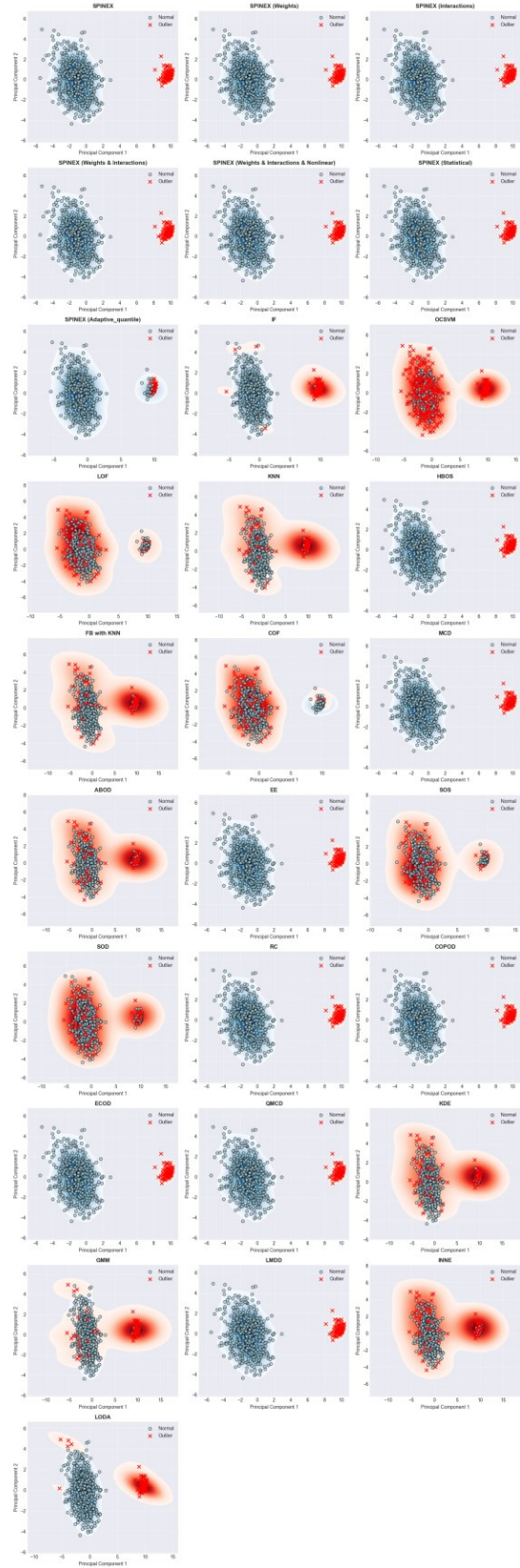
597
47

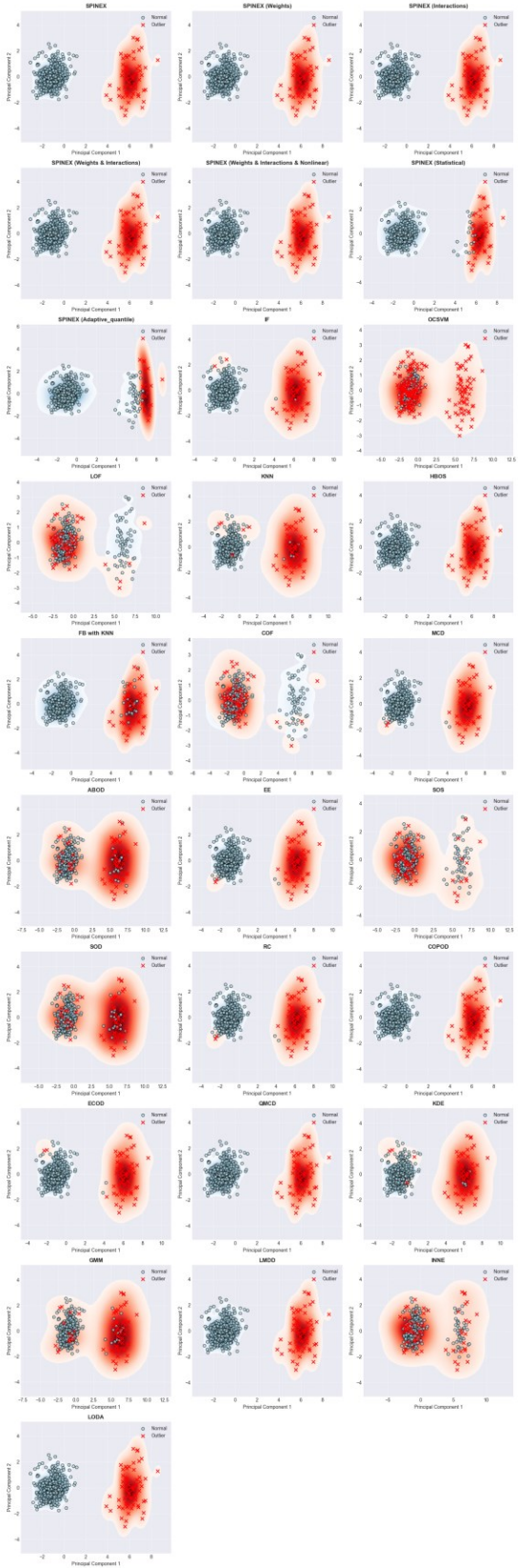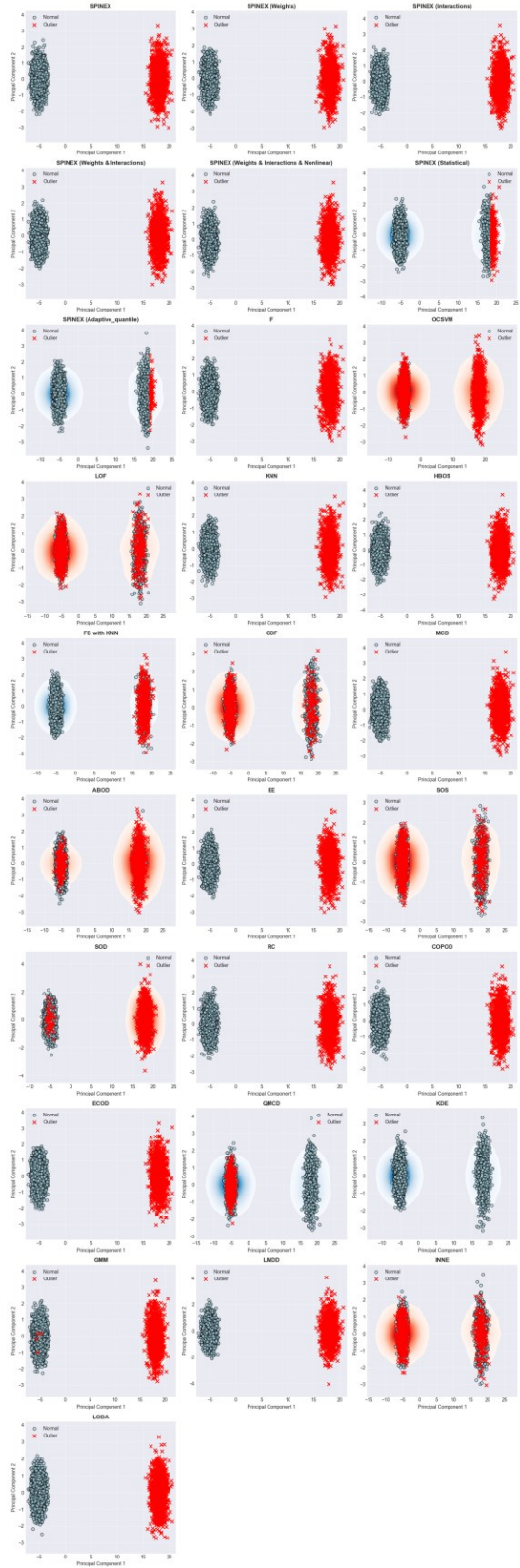

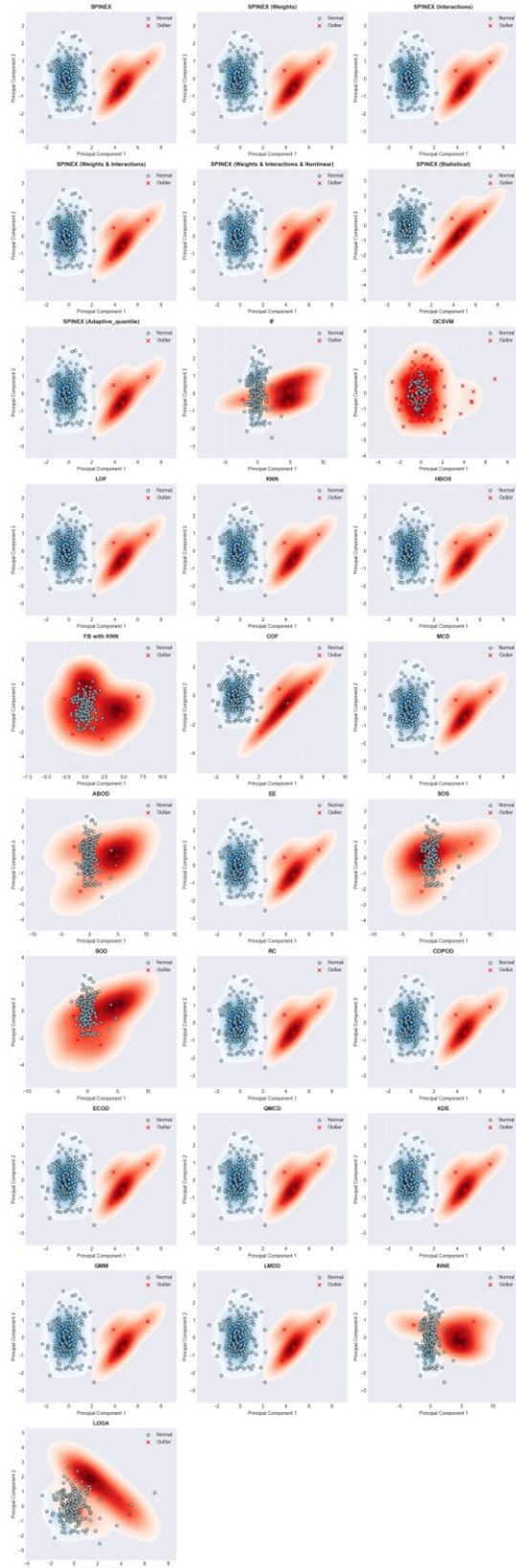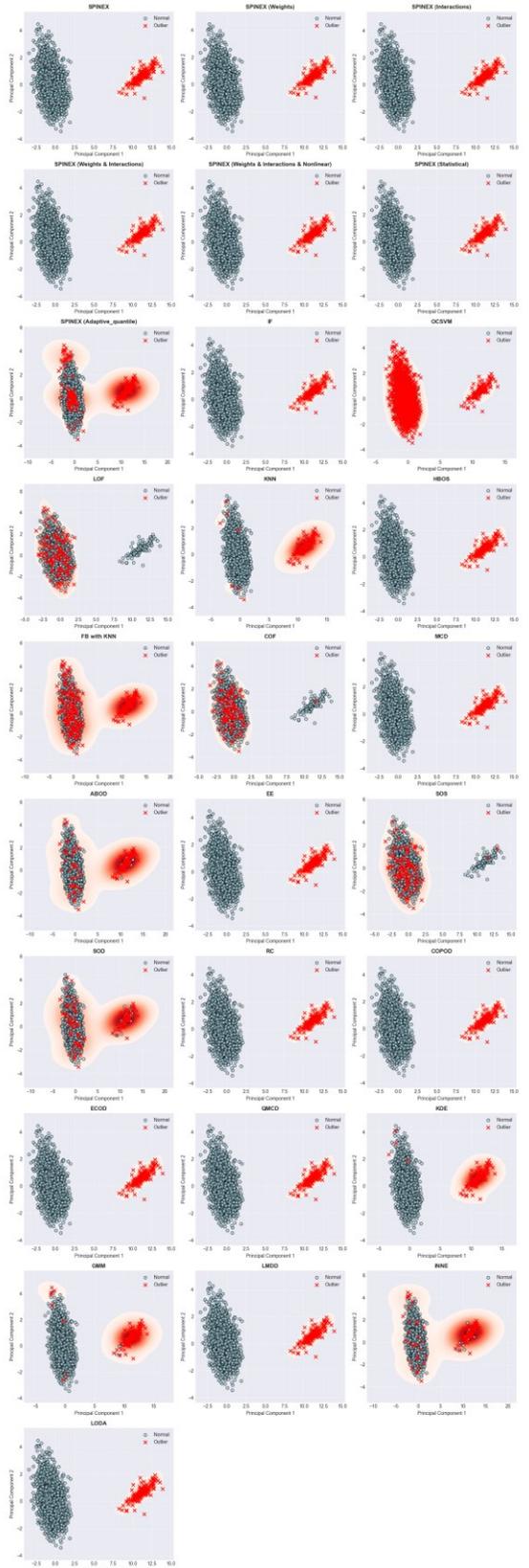

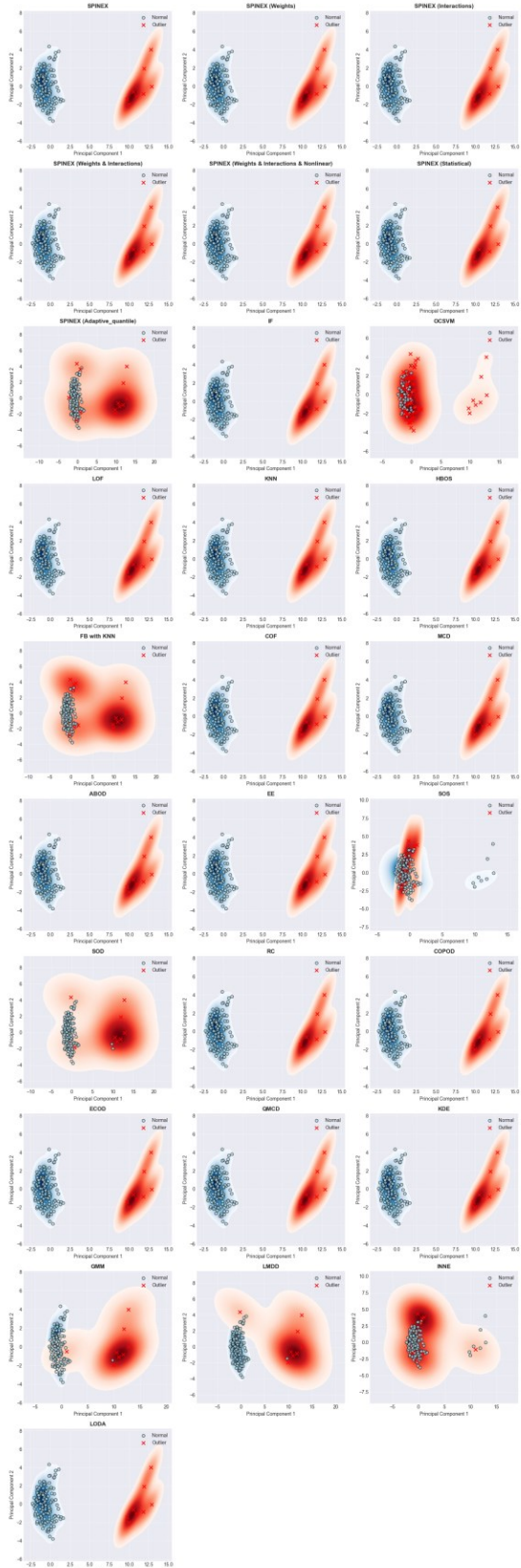
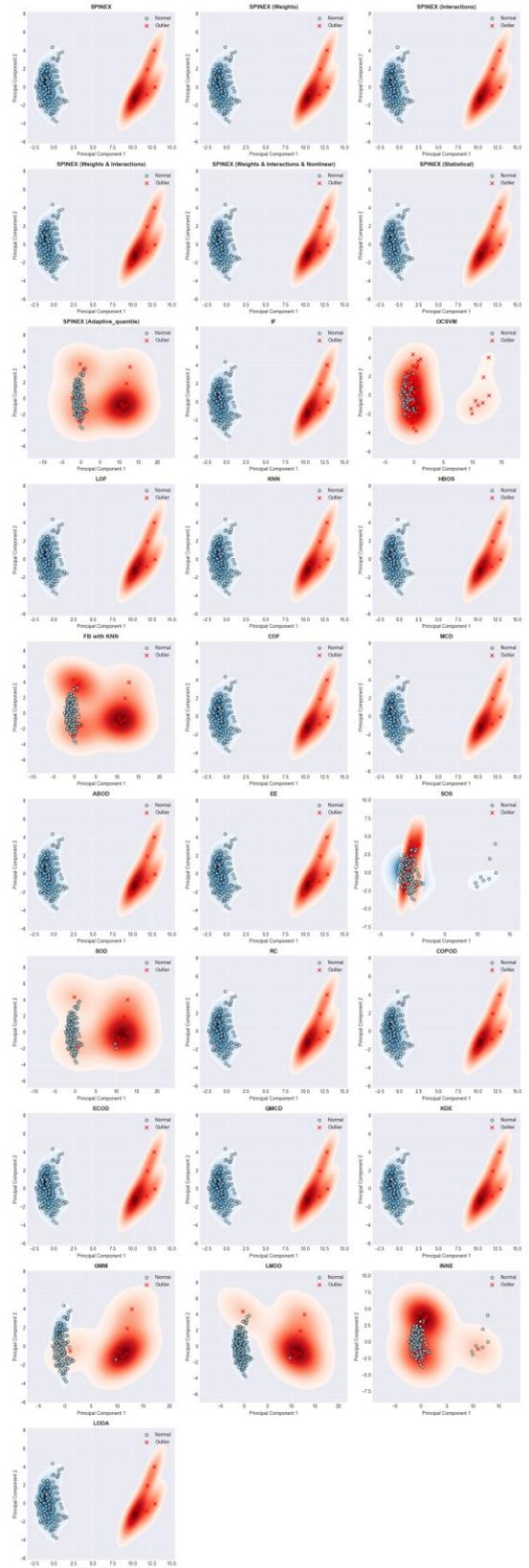

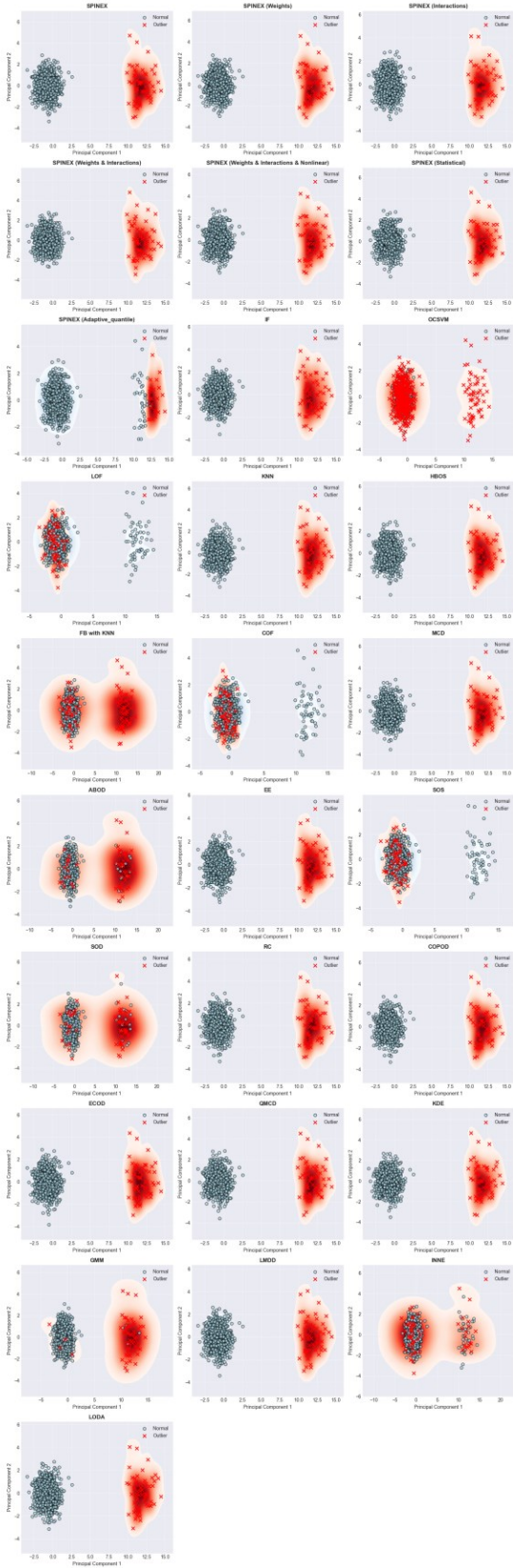

601

## C. Visualization of real datasets

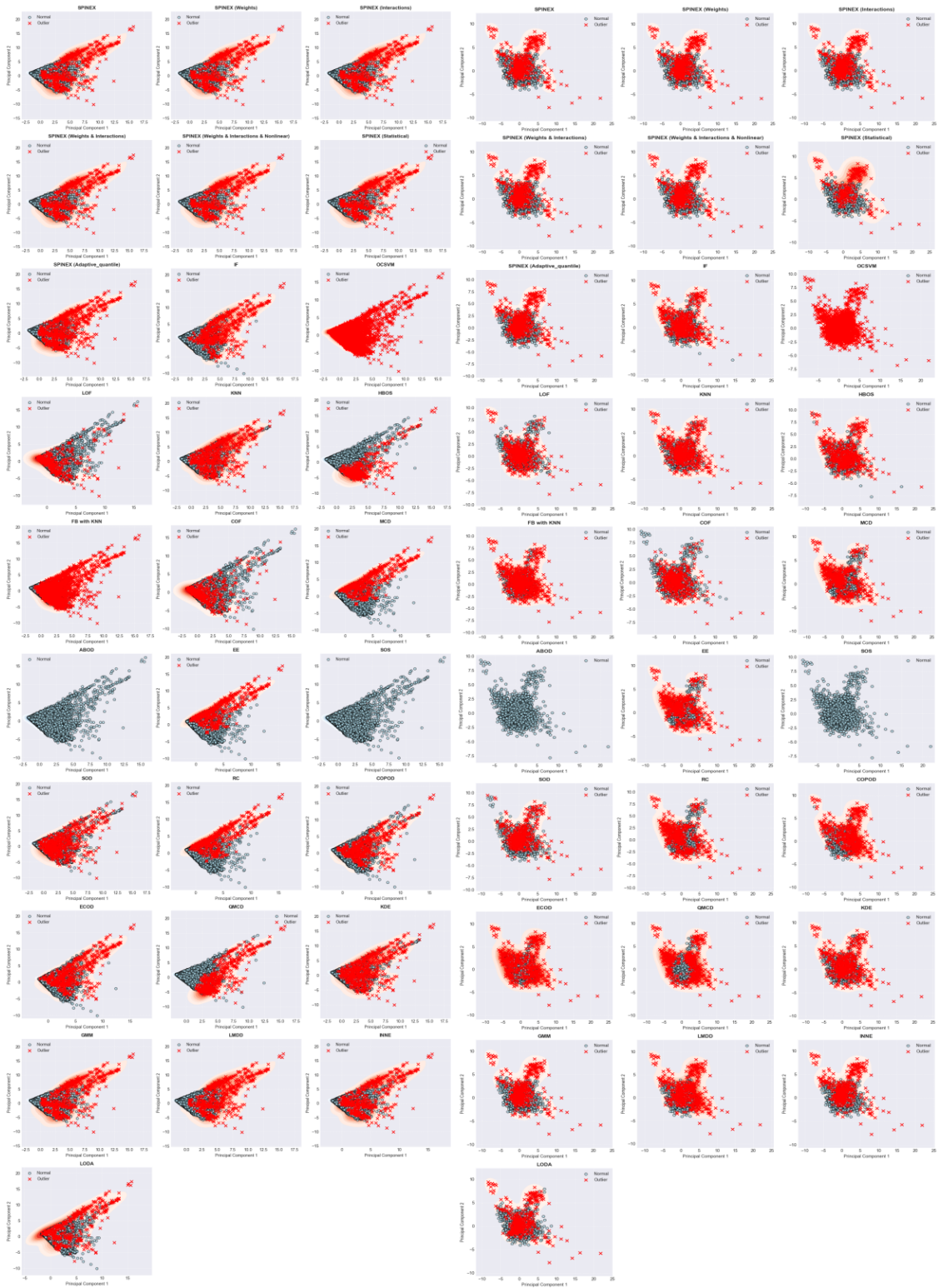

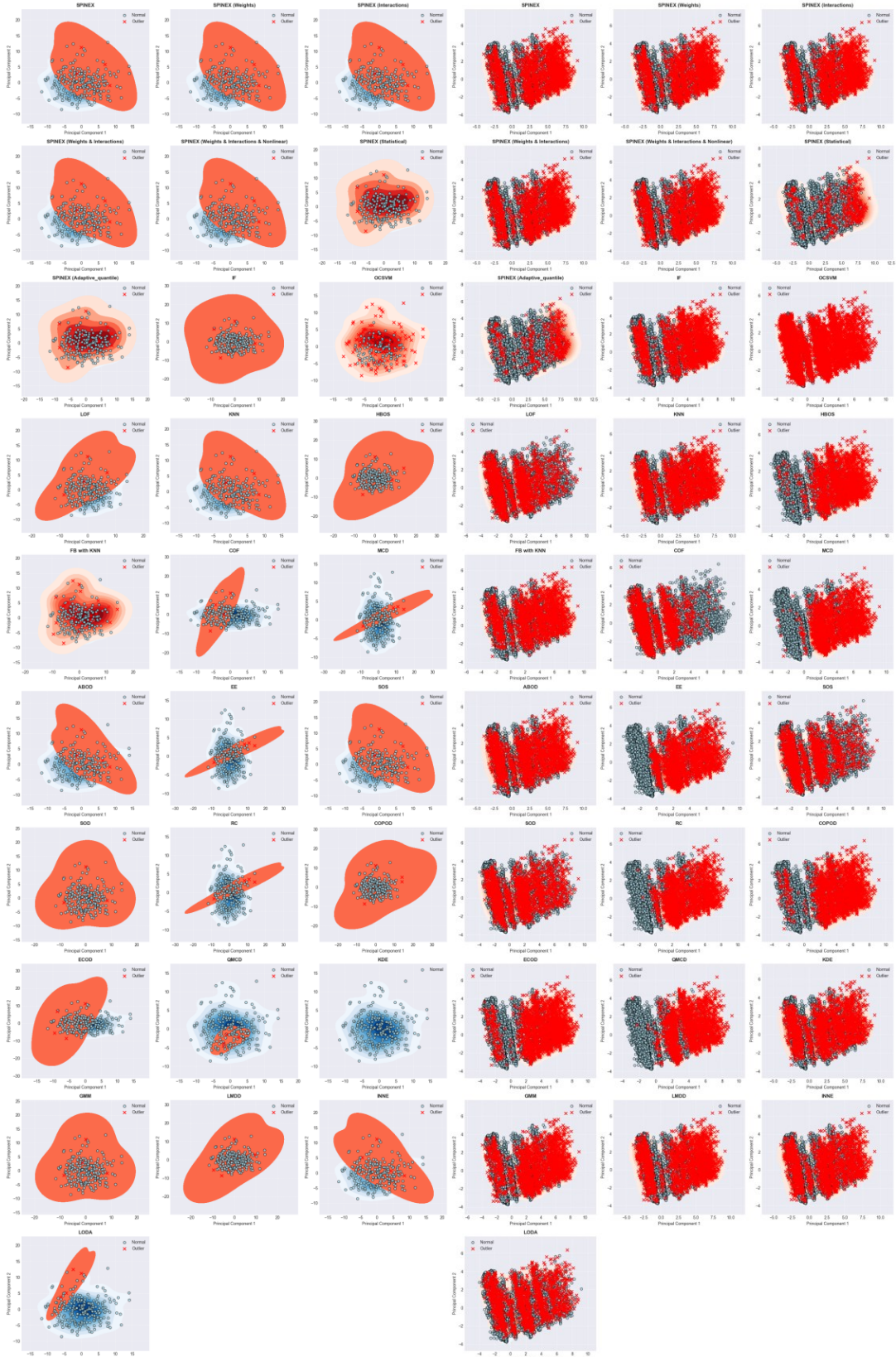

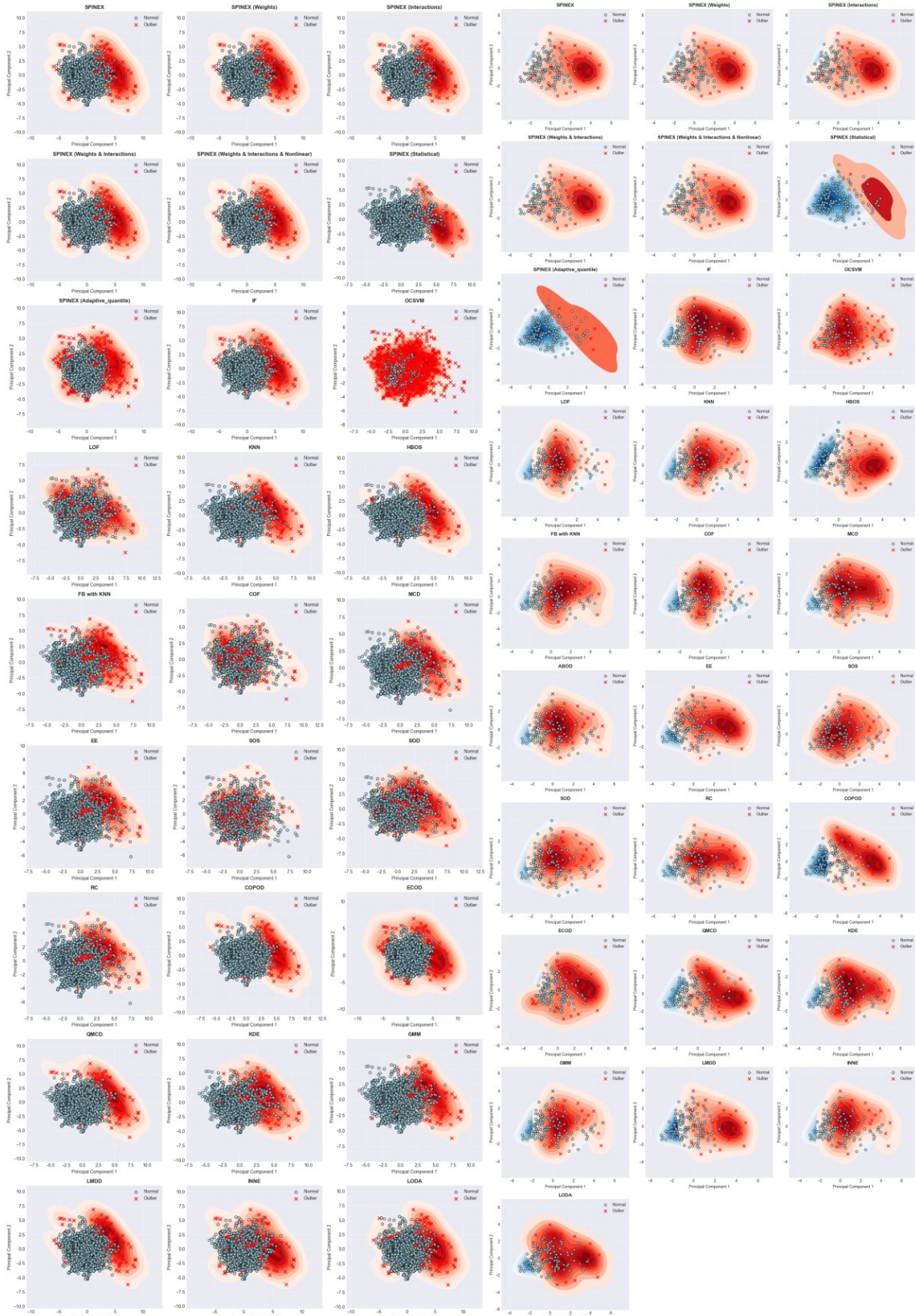

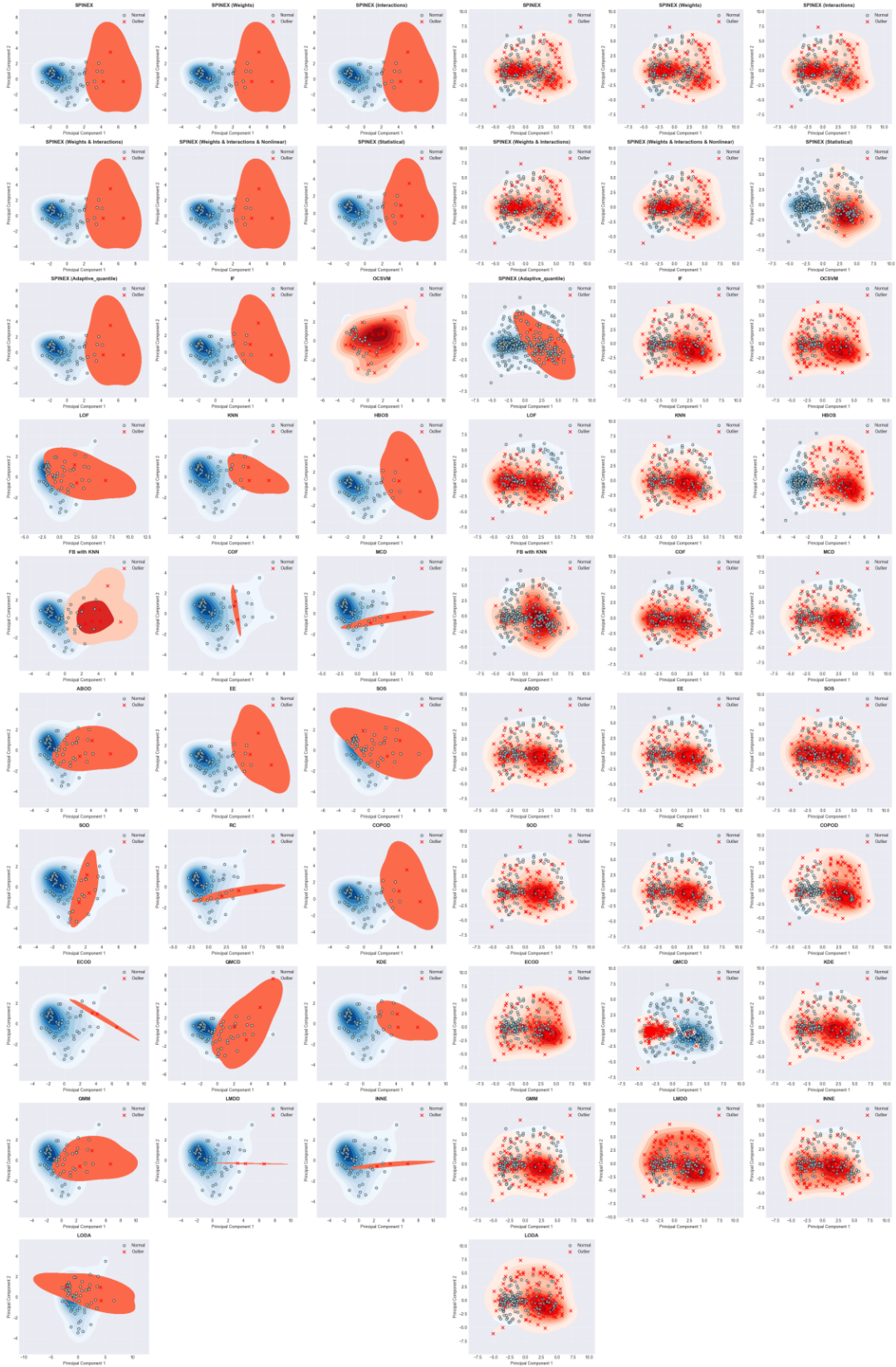

606

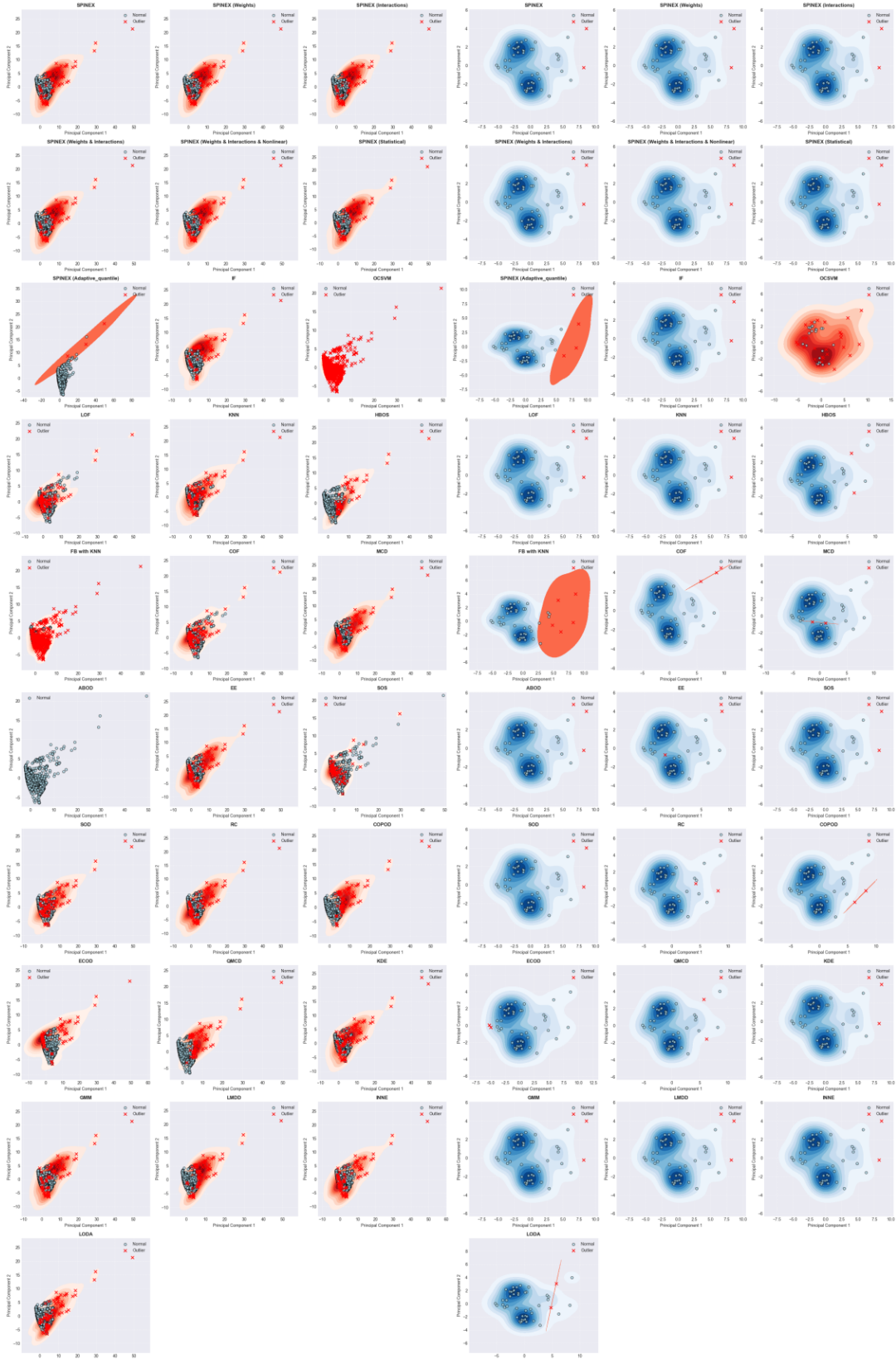

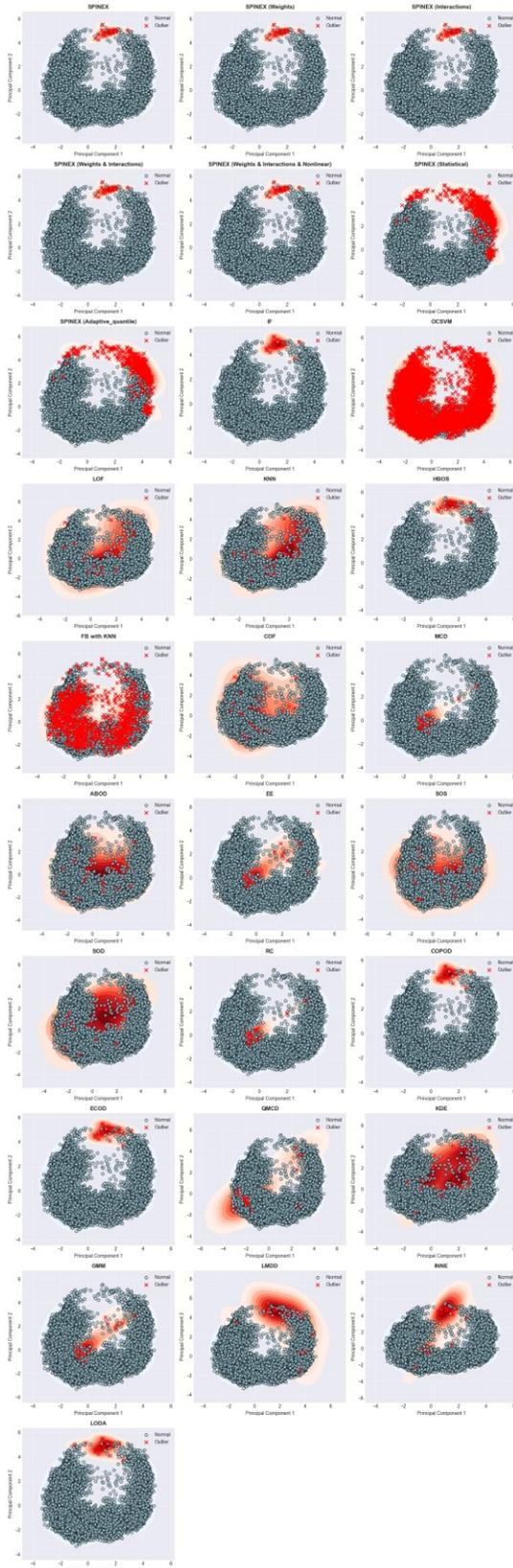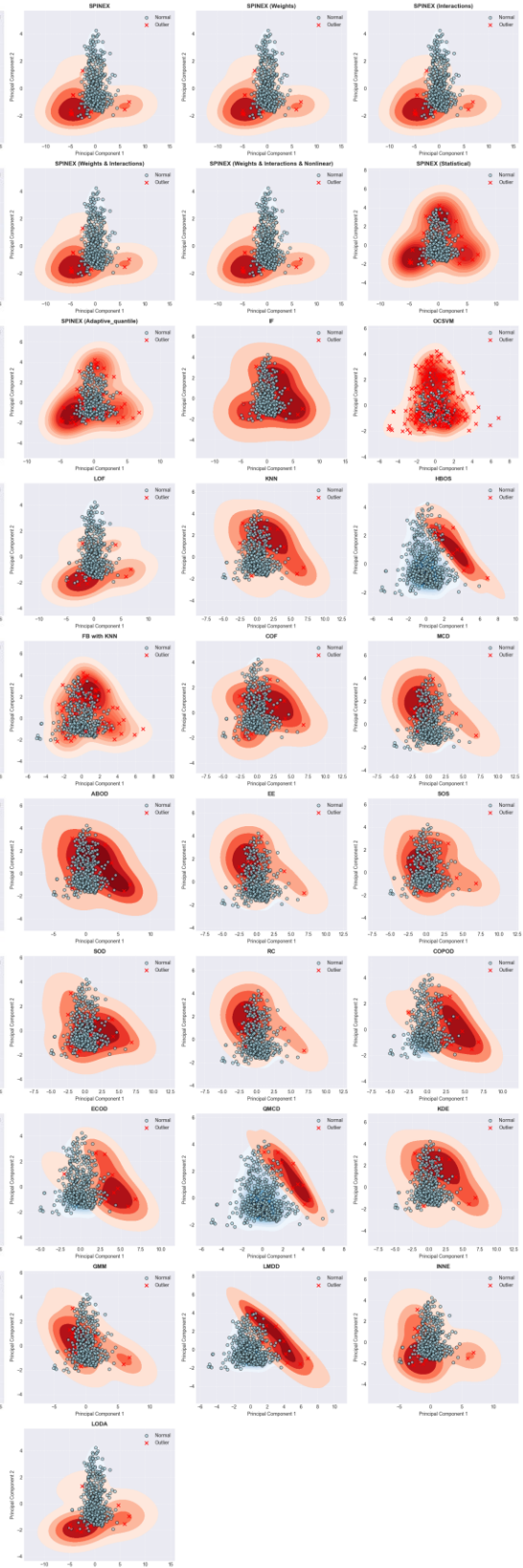608

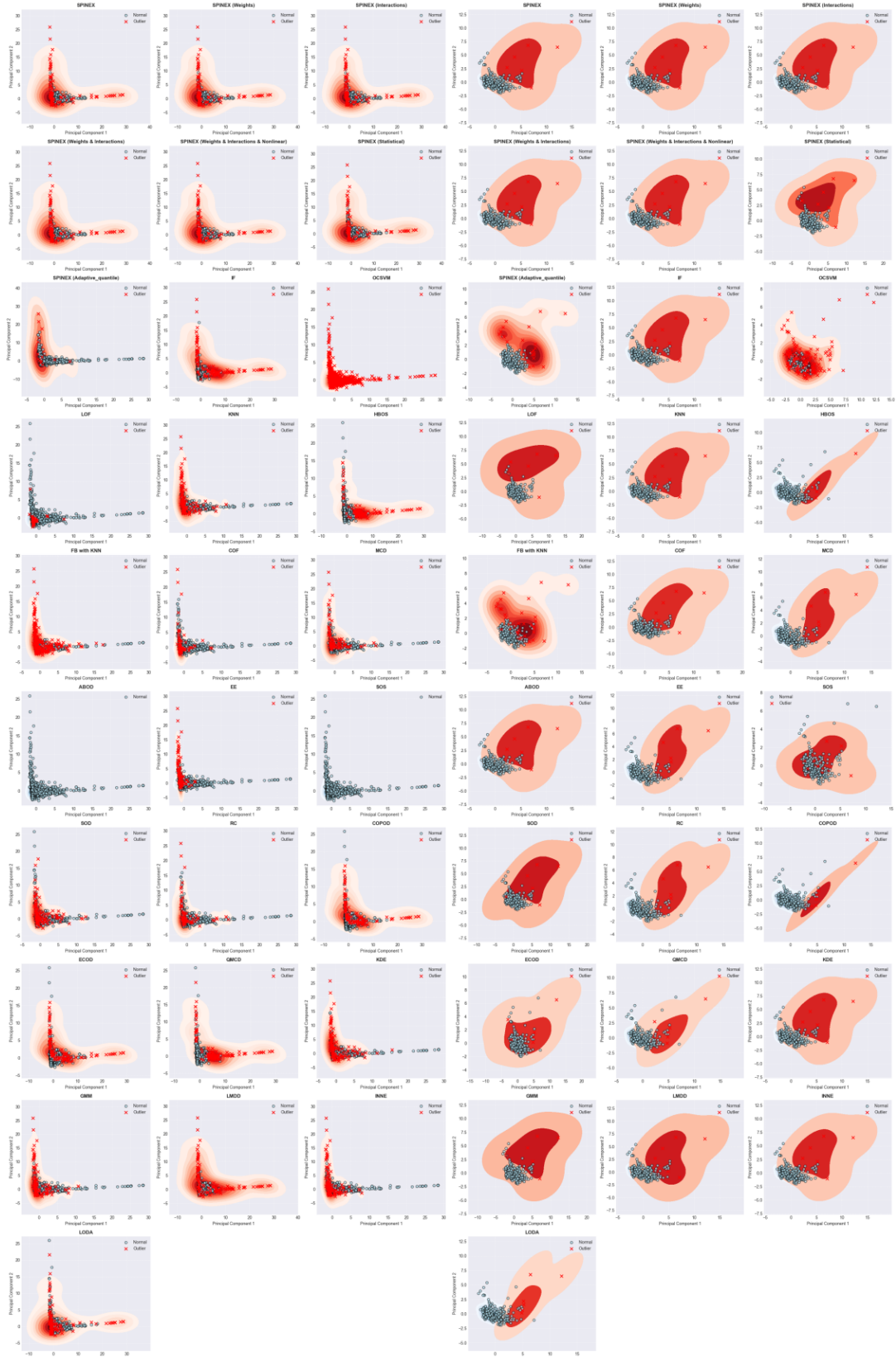

609

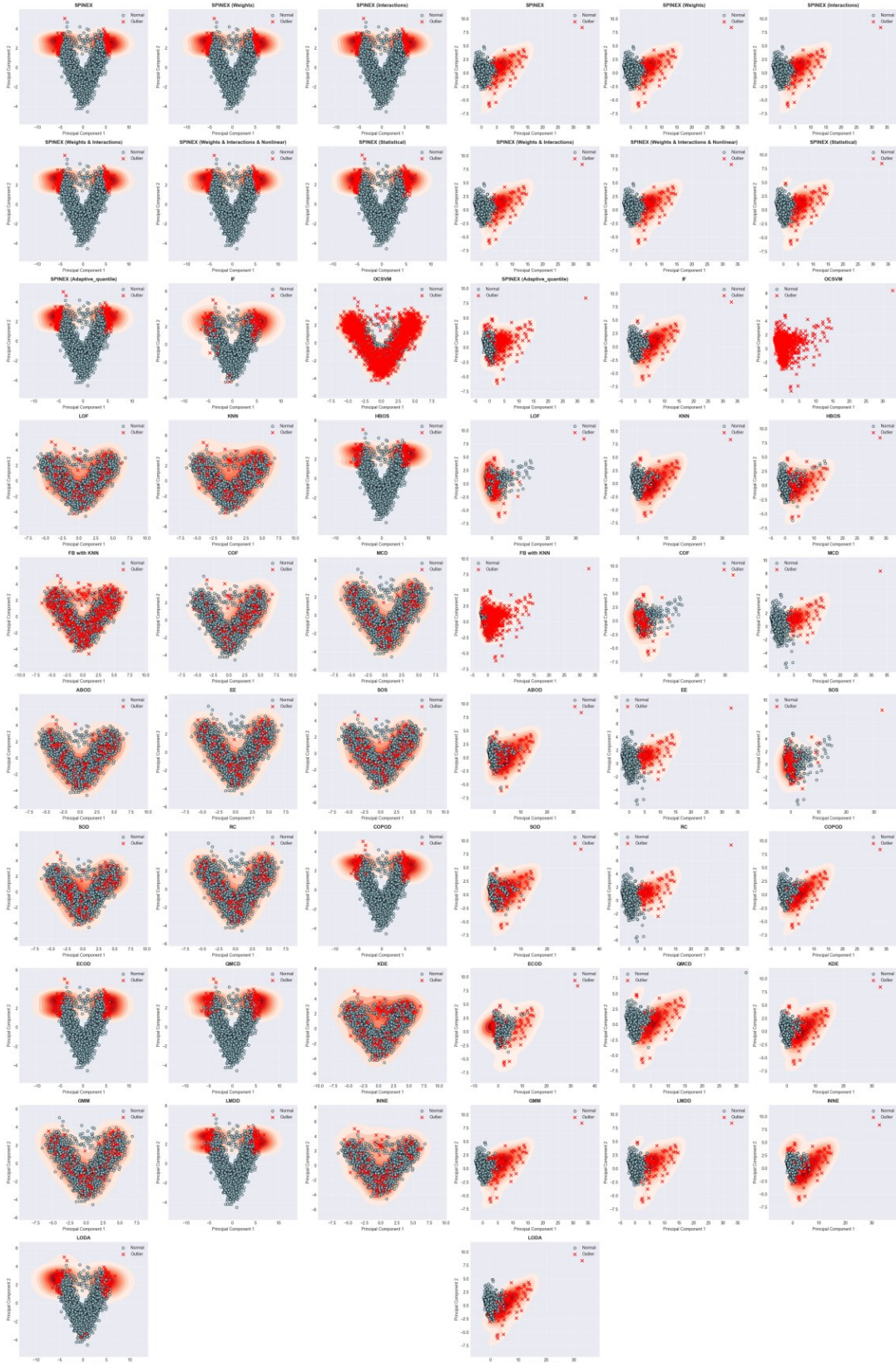

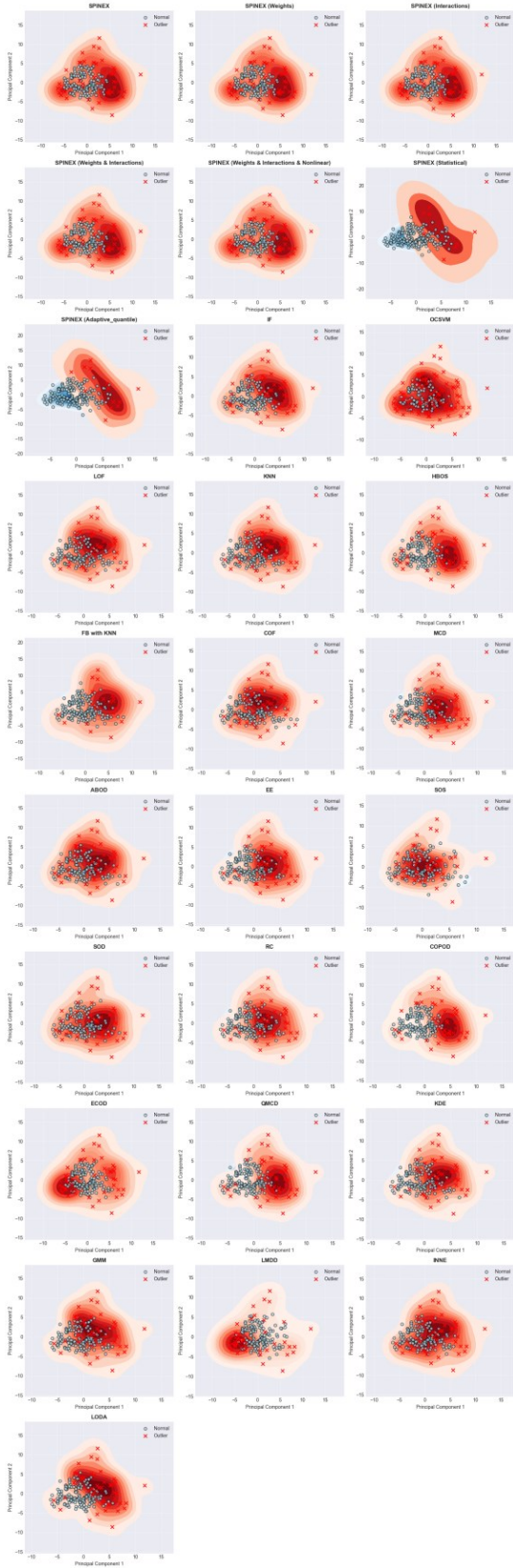

611

60